\documentclass[english,twoside,11pt]{article}
\usepackage[latin9]{inputenc}
\usepackage{color}
\usepackage{babel}
\usepackage{array}
\usepackage{float}
\usepackage{multirow}
\usepackage{amsmath}
\usepackage{amssymb}
\usepackage{graphicx}
\usepackage{xargs}[2008/03/08]
\usepackage[unicode=true,
 bookmarks=true,bookmarksnumbered=false,bookmarksopen=false,
 breaklinks=false,pdfborder={0 0 1},backref=false,colorlinks=true]
 {hyperref}
\usepackage{breakurl}

\makeatletter

\providecommand{\tabularnewline}{\\}
\floatstyle{ruled}
\newfloat{algorithm}{tbp}{loa}
\providecommand{\algorithmname}{Algorithm}
\floatname{algorithm}{\protect\algorithmname}

 \usepackage{algolyx}
 \usepackage{algolyx}

\usepackage{jair}
\usepackage{flushend}
\usepackage{theapa}
\usepackage{rawfonts}
\usepackage{caption}
\usepackage{algolyx}
\usepackage{algorithm}
\usepackage{algorithmic}
\usepackage{balance}
\usepackage{mathptmx}

\newfloat{algorithm}{tbp}{loa}[section]

\usepackage{colortbl} 
\definecolor{header_color}{rgb}{0.74,0.88,0.91}
\definecolor{even_color}{rgb}{0.9,0.9,0.9}
\definecolor{subheader_color}{rgb}{0.85,0.93,0.95}
\definecolor{childheader_color}{rgb}{1.0,0.93,0.87}

\@ifundefined{showcaptionsetup}{}{%
 \PassOptionsToPackage{caption=false}{subfig}}
\usepackage{subfig}
\makeatother

\begin{document}
\jairheading{1}{2017}{1-24}{xx/17}{xx/17} 

\ShortHeadings{Budgeted Batch Bayesian Optimization}{Nguyen, Rana, Gupta, Li, \& Venkatesh}

\title{Budgeted Batch Bayesian Optimization With Unknown Batch Sizes}

\author{\name Vu Nguyen \email v.nguyen@deakin.edu.au \\        \name Santu Rana \email santu.rana@deakin.edu.au \\  \name Sunil Gupta \email sunil.gupta@deakin.edu.au \\ \name Cheng Li \email cheng.l@deakin.edu.au \\ \name Svetha Venkatesh \email svetha.venkatesh@deakin.edu.au \\       \addr Center for Pattern Recognition and Data Analytics\\        Deakin University, Geelong, Australia      }

\maketitle
\begin{abstract}
Parameter settings profoundly impact the performance of machine learning
algorithms and laboratory experiments. The classical grid search or
trial-error methods are exponentially expensive in large parameter
spaces, and Bayesian optimization (BO) offers an elegant alternative
for global optimization of black box functions. In situations where
the black box function can be evaluated at multiple points simultaneously,
batch Bayesian optimization is used. Current batch BO approaches are
restrictive in that they fix the number of evaluations per batch,
and this can be wasteful when the number of specified evaluations
is larger than the number of real maxima in the underlying acquisition
function. We present the Budgeted Batch Bayesian Optimization (B3O)
for hyper-parameter tuning and experimental design - we identify the
appropriate batch size for each iteration in an elegant way. To set
the batch size flexible, we use the infinite Gaussian mixture model
(IGMM) for automatically identifying the number of peaks in the underlying
acquisition functions. We solve the intractability of estimating the
IGMM directly from the acquisition function by formulating the batch
generalized slice sampling to efficiently draw samples from the acquisition
function. We perform extensive experiments for both synthetic functions
and two real world applications - machine learning hyper-parameter
tuning and experimental design for alloy hardening. We show empirically
that the proposed B3O outperforms the existing fixed batch BO approaches
in finding the optimum whilst requiring a fewer number of evaluations,
thus saving cost and time.
\end{abstract}
\newcommand{\sidenote}[1]{\marginpar{\small \emph{\color{Medium}#1}}}

\global\long\def\se{\hat{\text{se}}}

\global\long\def\interior{\text{int}}

\global\long\def\boundary{\text{bd}}

\global\long\def\ML{\textsf{ML}}

\global\long\def\GML{\mathsf{GML}}

\global\long\def\HMM{\mathsf{HMM}}

\global\long\def\support{\text{supp}}

\global\long\def\new{\text{*}}

\global\long\def\stir{\text{Stirl}}

\global\long\def\mA{\mathcal{A}}

\global\long\def\mB{\mathcal{B}}

\global\long\def\mF{\mathcal{F}}

\global\long\def\mK{\mathcal{K}}

\global\long\def\mH{\mathcal{H}}

\global\long\def\mX{\mathcal{X}}

\global\long\def\mZ{\mathcal{Z}}

\global\long\def\mS{\mathcal{S}}

\global\long\def\Ical{\mathcal{I}}

\global\long\def\mT{\mathcal{T}}

\global\long\def\Pcal{\mathcal{P}}

\global\long\def\dist{d}

\global\long\def\HX{\entro\left(X\right)}
 \global\long\def\entropyX{\HX}

\global\long\def\HY{\entro\left(Y\right)}
 \global\long\def\entropyY{\HY}

\global\long\def\HXY{\entro\left(X,Y\right)}
 \global\long\def\entropyXY{\HXY}

\global\long\def\mutualXY{\mutual\left(X;Y\right)}
 \global\long\def\mutinfoXY{\mutualXY}

\global\long\def\gv{\given}

\global\long\def\goto{\rightarrow}

\global\long\def\asgoto{\stackrel{a.s.}{\longrightarrow}}

\global\long\def\pgoto{\stackrel{p}{\longrightarrow}}

\global\long\def\dgoto{\stackrel{d}{\longrightarrow}}

\global\long\def\lik{\mathcal{L}}

\global\long\def\logll{\mathit{l}}

\global\long\def\vectorize#1{\mathbf{#1}}

\global\long\def\vt#1{\mathbf{#1}}

\global\long\def\gvt#1{\boldsymbol{#1}}

\global\long\def\idp{\ \bot\negthickspace\negthickspace\bot\ }
 \global\long\def\cdp{\idp}

\global\long\def\das{}

\global\long\def\id{\mathbb{I}}

\global\long\def\idarg#1#2{\id\left\{  #1,#2\right\}  }

\global\long\def\iid{\stackrel{\text{iid}}{\sim}}

\global\long\def\bzero{\vt 0}

\global\long\def\bone{\mathbf{1}}

\global\long\def\boldm{\boldsymbol{m}}

\global\long\def\bff{\vt f}

\global\long\def\bx{\boldsymbol{x}}

\global\long\def\bc{\boldsymbol{c}}

\global\long\def\bl{\boldsymbol{l}}

\global\long\def\bu{\boldsymbol{u}}

\global\long\def\bo{\boldsymbol{o}}

\global\long\def\bk{\boldsymbol{k}}

\global\long\def\bh{\boldsymbol{h}}

\global\long\def\bs{\boldsymbol{s}}

\global\long\def\bz{\boldsymbol{z}}

\global\long\def\xnew{y}

\global\long\def\bxnew{\boldsymbol{y}}

\global\long\def\bX{\boldsymbol{X}}

\global\long\def\tbx{\tilde{\bx}}

\global\long\def\by{\boldsymbol{y}}

\global\long\def\bY{\boldsymbol{Y}}

\global\long\def\bZ{\boldsymbol{Z}}

\global\long\def\bU{\boldsymbol{U}}

\global\long\def\bv{\boldsymbol{v}}

\global\long\def\bn{\boldsymbol{n}}

\global\long\def\bV{\boldsymbol{V}}

\global\long\def\bK{\boldsymbol{K}}

\global\long\def\bI{\boldsymbol{I}}

\global\long\def\bw{\vt w}

\global\long\def\bbeta{\gvt{\beta}}

\global\long\def\bmu{\gvt{\mu}}

\global\long\def\btheta{\boldsymbol{\theta}}

\global\long\def\blambda{\boldsymbol{\lambda}}

\global\long\def\bgamma{\boldsymbol{\gamma}}

\global\long\def\bpsi{\boldsymbol{\psi}}

\global\long\def\bphi{\boldsymbol{\phi}}

\global\long\def\bpi{\boldsymbol{\pi}}

\global\long\def\eeta{\boldsymbol{\eta}}

\global\long\def\bomega{\boldsymbol{\omega}}

\global\long\def\bepsilon{\boldsymbol{\epsilon}}

\global\long\def\btau{\boldsymbol{\tau}}

\global\long\def\bSigma{\gvt{\Sigma}}

\global\long\def\realset{\mathbb{R}}

\global\long\def\realn{\realset^{n}}

\global\long\def\integerset{\mathbb{Z}}

\global\long\def\natset{\integerset}

\global\long\def\integer{\integerset}

\global\long\def\natn{\natset^{n}}

\global\long\def\rational{\mathbb{Q}}

\global\long\def\rationaln{\rational^{n}}

\global\long\def\complexset{\mathbb{C}}

\global\long\def\comp{\complexset}

\global\long\def\compl#1{#1^{\text{c}}}

\global\long\def\and{\cap}

\global\long\def\compn{\comp^{n}}

\global\long\def\comb#1#2{\left({#1\atop #2}\right) }

\global\long\def\nchoosek#1#2{\left({#1\atop #2}\right)}

\global\long\def\param{\vt w}

\global\long\def\Param{\Theta}

\global\long\def\meanparam{\gvt{\mu}}

\global\long\def\Meanparam{\mathcal{M}}

\global\long\def\meanmap{\mathbf{m}}

\global\long\def\logpart{A}

\global\long\def\simplex{\Delta}

\global\long\def\simplexn{\simplex^{n}}

\global\long\def\dirproc{\text{DP}}

\global\long\def\ggproc{\text{GG}}

\global\long\def\DP{\text{DP}}

\global\long\def\ndp{\text{nDP}}

\global\long\def\hdp{\text{HDP}}

\global\long\def\gempdf{\text{GEM}}

\global\long\def\ei{\text{EI}}

\global\long\def\rfs{\text{RFS}}

\global\long\def\bernrfs{\text{BernoulliRFS}}

\global\long\def\poissrfs{\text{PoissonRFS}}

\global\long\def\grad{\gradient}
 \global\long\def\gradient{\nabla}

\global\long\def\cpr#1#2{\Pr\left(#1\ |\ #2\right)}

\global\long\def\var{\text{Var}}

\global\long\def\Var#1{\text{Var}\left[#1\right]}

\global\long\def\cov{\text{Cov}}

\global\long\def\Cov#1{\cov\left[ #1 \right]}

\global\long\def\COV#1#2{\underset{#2}{\cov}\left[ #1 \right]}

\global\long\def\corr{\text{Corr}}

\global\long\def\sst{\text{T}}

\global\long\def\SST{\sst}

\global\long\def\ess{\mathbb{E}}

\global\long\def\Ess#1{\ess\left[#1\right]}

\newcommandx\ESS[2][usedefault, addprefix=\global, 1=]{\underset{#2}{\ess}\left[#1\right]}

\global\long\def\fisher{\mathcal{F}}

\global\long\def\bfield{\mathcal{B}}
 \global\long\def\borel{\mathcal{B}}

\global\long\def\bernpdf{\text{Bernoulli}}

\global\long\def\betapdf{\text{Beta}}

\global\long\def\dirpdf{\text{Dir}}

\global\long\def\gammapdf{\text{Gamma}}

\global\long\def\gaussden#1#2{\text{Normal}\left(#1, #2 \right) }

\global\long\def\gauss{\mathbf{N}}

\global\long\def\gausspdf#1#2#3{\text{Normal}\left( #1 \lcabra{#2, #3}\right) }

\global\long\def\multpdf{\text{Mult}}

\global\long\def\poiss{\text{Pois}}

\global\long\def\poissonpdf{\text{Poisson}}

\global\long\def\pgpdf{\text{PG}}

\global\long\def\wshpdf{\text{Wish}}

\global\long\def\iwshpdf{\text{InvWish}}

\global\long\def\nwpdf{\text{NW}}

\global\long\def\niwpdf{\text{NIW}}

\global\long\def\studentpdf{\text{Student}}

\global\long\def\unipdf{\text{Uni}}

\global\long\def\transp#1{\transpose{#1}}
 \global\long\def\transpose#1{#1^{\mathsf{T}}}

\global\long\def\mgt{\succ}

\global\long\def\mge{\succeq}

\global\long\def\idenmat{\mathbf{I}}

\global\long\def\trace{\mathrm{tr}}

\global\long\def\argmax#1{\underset{_{#1}}{\text{argmax}} }

\global\long\def\argmin#1{\underset{_{#1}}{\text{argmin}\ } }

\global\long\def\diag{\text{diag}}

\global\long\def\norm{}

\global\long\def\spn{\text{span}}

\global\long\def\vtspace{\mathcal{V}}

\global\long\def\field{\mathcal{F}}
 \global\long\def\ffield{\mathcal{F}}

\global\long\def\inner#1#2{\left\langle #1,#2\right\rangle }
 \global\long\def\iprod#1#2{\inner{#1}{#2}}

\global\long\def\dprod#1#2{#1 \cdot#2}

\global\long\def\norm#1{\left\Vert #1\right\Vert }

\global\long\def\entro{\mathbb{H}}

\global\long\def\entropy{\mathbb{H}}

\global\long\def\Entro#1{\entro\left[#1\right]}

\global\long\def\Entropy#1{\Entro{#1}}

\global\long\def\mutinfo{\mathbb{I}}

\global\long\def\relH{\mathit{D}}

\global\long\def\reldiv#1#2{\relH\left(#1||#2\right)}

\global\long\def\KL{KL}

\global\long\def\KLdiv#1#2{\KL\left(#1\parallel#2\right)}
 \global\long\def\KLdivergence#1#2{\KL\left(#1\ \parallel\ #2\right)}

\global\long\def\crossH{\mathcal{C}}
 \global\long\def\crossentropy{\mathcal{C}}

\global\long\def\crossHxy#1#2{\crossentropy\left(#1\parallel#2\right)}

\global\long\def\breg{\text{BD}}

\global\long\def\lcabra#1{\left|#1\right.}

\global\long\def\lbra#1{\lcabra{#1}}

\global\long\def\rcabra#1{\left.#1\right|}

\global\long\def\rbra#1{\rcabra{#1}}

\section{Introduction}

Global optimization is fundamental to diverse real-world problems
where parameter settings and design choices are pivotal - as an example,
in algorithm performance (deep learning networks \cite{Bengio_09learning})
or quality of the products (chemical processes or engineering design
\cite{Wang_2007Review}). This requires us to find the global maximum
of a non-concave objective function using sequential, and often, noisy
observations. Critically, the objective functions are unknown and
expensive to evaluate. Therefore, the challenge is to find the maximum
of such expensive objective functions in few sequential queries, thus
minimizing time and cost. 

Bayesian optimization (BO) is an approach to find the global optimum
of such expensive, black-box objective functions $f$ using limited
evaluations \cite{Snoek_2012Practical,Shahriari_2016Taking,Wang_2016Bayesian}.
Instead of optimizing the real function, BO utilizes a cheaper to
evaluate surrogate function, called the acquisition function. This
function is used to suggest the next point by balancing exploitation
(knowledge of what has been observed) and exploration (where a function
has not been investigated). Then, after evaluating the objective function
at the suggested point, a Gaussian process \cite{Rasmussen_2006gaussian}
is updated and thus gradually a profile of the mean and uncertainty
of the exploration space is built up. Bayesian optimization has been
demonstrated to outperform other state-of-the-art black-box optimization
techniques when function evaluations are expensive and the number
of allowed function evaluations is low \cite{Hutter_2013Evaluation}.
Thus, BO has received increasing attention in machine learning community
\cite{Thornton_2013Auto,Li_NIPSW2016High,Khajah_2016Designing}. 

Most selection strategies in BO are sequential, wherein only one experiment
is tested at a time - the experiment selection at each iteration is
optimized by using the available observed information. However, such
methods are inefficient when parallel evaluations are possible. For
examples, many alloy samples can be placed in an oven simultaneously
(for testing the alloy quality), or several machine learning algorithms
can be run in parallel at the same time using multiple cores. This
motivates batch Bayesian optimization algorithms that select multiple
experiments, a batch of \emph{$q$ }experiments, at each iteration
for evaluation. 

Existing batch BO approaches are mostly greedy, sequentially visiting
all the maxima of the acquisition function. After the first maximum
is found, such methods modify the acquisition function by suppressing
the current maximum and then move on to find the next best maximum.
This repeats till a batch of maxima is collected. Different algorithms
implement this philosophy -\cite{Ginsbourger_2008Multi} modifies
the found maximum point by replacing it with a ``fake'' or constant
value, thus the algorithm is called ``constant liar''. Other approaches,
including GP-BUCB \cite{Desautels_2014Parallelizing} and the GP-UCB-PE
\cite{Contal_2013Parallel}, exploit the predictive variance of GPs
that only depend on the features $\bx$, but not the outcome values
$\by$. Local Penalization \cite{Gonzalez_2015Batch}, on the other
hand, using the estimation of Lipschitz constant to penalize the peaks.
All these methods update the posterior variance sequentially to modify
the acquisition function and thereby derive a batch. Most importantly,
these batch BO methods are essentially greedy, choosing individual
points until the batch is filled. This is not optimal as after the
``real'' maxima in the function are found, noisy points will get
added simply to complete the batch. An alternate and non-greedy approach
is to select a batch of points that reduce the uncertainty of the
global maximizer \cite{Shah_2015Parallel} by extending the Predictive
Entropy Search (PES) algorithm \cite{Hernandez_2014Predictive} for
a batch setting. But again, a fixed batch size is used. Fixed batch
size approaches can restrict the flexibility in finding the global
solution and require unnecessary evaluations or miss evaluation of
important points. If we over-specify the batch size, we waste resources
in evaluating redundant points. In contrast, if we under-specify the
batch size, we may miss important points that could potentially be
the optimal solution of $f$.

The best solution is to fill each batch \emph{flexibly}, but the challenge
is to know exactly \emph{how many maxima are there in each batch}.
Our key contribution is to solve this problem, by estimating the number
of maxima in the acquisition function, we let the algorithm adjust
the batch size at each iteration. This flexible and non-greedy approach
has the advantage of being efficient in that it suggests only as many
maxima locations as needed. The solution comes from our intuition
that the multiple peaks in the acquisition function can be approximated
by a mixture of Gaussians - the means of the Gaussian correspond to
the underlying peaks. However, since the number of underlying peaks
is unknown, we use the infinite Gaussian mixture model (IGMM) \cite{Rasmussen_1999_Infinite}
so that the number of peaks in the acquisition function can be estimated
automatically. Because fitting the IGMM directly to the acquisition
function is intractable, we present an efficient batch generalized
slice sampler approach to draw samples from acquisition function which
are then used to learn the unknown number of peaks. Although the IGMM
and the slice sampler have existed for more than a decade, the idea
of making use of these techniques for batch Bayesian optimization
is novel. 

In the experiments, we first validate our methods in 8 benchmark functions
and compare it against 9 baselines. We use several criteria for comparison:
a) in \emph{best-found-value}, our method achieves better results
in 5 out of 8 cases; and b) our method matches the baseline performance
but requires far \emph{fewer evaluation points}. We further demonstrate
the algorithm on machine learning hyper-parameter tuning for support
vector regression, multi-label classification and deep learning. We
show that our model outperforms the baselines in finding the best
values (RMSE, F1 score and accuracy) whilst we require fewer experimental
evaluations to reach targets. In addition, we perform a real world
experimental design in which we formulate the heat-treatment process
required for producing an Aluminum-scandium alloy. We show that our
method produces the highest hardness using the fewest number of evaluations.
Our result is significant as searching for the best heat-treatment
process is costly and making this process efficient results in cost
and time savings.

\section{Preliminary}

We first review the Gaussian process (GP), Bayesian optimization (BO)
and the acquisition functions. Then, we summarize the batch Bayesian
optimization setting and existing approaches.

\subsection{Gaussian process}

Gaussian processes (GP) \cite{Rasmussen_2006gaussian} extends a multivariate
Gaussian distribution to infinite dimensionality. Formally, Gaussian
process generates data located throughout some domains such that any
finite subset of the range follows a multivariate Gaussian distribution.
Given $N$ observations $\bY=\left\{ y_{1},y_{2},...y_{N}\right\} $
which can always be imagined as a single point sampled from some multivariate
Gaussian distributions.

The mean of GP is assumed to be zero everywhere. What relates one
observation to another in such cases is just the covariance function,
$k\left(x,x'\right)$. From the assumption of GP, we have $\by\sim\mathcal{N}\left(0,\bK\right)$
where the covariance matrix is defined as follows:
\[
\bK=\left[\begin{array}{cccc}
k\left(x_{1},x_{1}\right) & k\left(x_{2},x_{2}\right) & \cdots & k\left(x_{1},x_{N}\right)\\
k\left(x_{2},x_{1}\right) & k\left(x_{2},x_{2}\right) & \cdots & k\left(x_{2},x_{N}\right)\\
\vdots & \vdots & \ddots & \vdots\\
k\left(x_{N},x_{1}\right) & k\left(x_{N},x_{2}\right) & \cdots & k\left(x_{N},x_{N}\right).
\end{array}\right]
\]
A popular choice for the covariance function $k$ is the squared exponential
function: $k\left(x,x'\right)=\sigma_{f}^{2}\exp\left[\frac{-\left(x-x'\right)^{2}}{2l^{2}}\right]$
where $\sigma_{f}^{2}$ defines the maximum allowable covariance.
It $x\thickapprox x'$, then $k\left(x,x'\right)$ approaches this
maximum of $\sigma_{f}^{2}$, indicating that $f(x)$ is perfectly
correlated with $f(x')$. If $x$ is far from $x'$, we have instead
$k\left(x,x'\right)\thickapprox0$. The length parameter $l$ will
control this separation when $x$ is not closed to $x'$.

For prediction on a new data point $y_{*}$, we can update the covariance
matrix with $k_{**}=k\left(x_{*},x_{*}\right)$ and $\bk_{*}=\begin{array}{cccc}
k\left(x_{*},x_{1}\right) & k\left(x_{*},x_{2}\right) & \cdots & k\left(x_{*},x_{N}\right)\end{array}$ . Hence, we can write
\begin{align*}
\left[\begin{array}{c}
\by\\
y_{*}
\end{array}\right] & \sim\mathcal{N}\left(0,\left[\begin{array}{cc}
\bK & \bk_{*}^{T}\\
\bk_{*} & k_{**}
\end{array}\right]\right).
\end{align*}
The conditional probability is followed Gaussian distribution as
$p\left(y_{*}\mid\by\right)\sim\mathcal{N}\left(\mu\left(\bx_{*}\right),\sigma^{2}\left(\bx_{*}\right)\right)$
\cite{Rasmussen_2006gaussian,Ebden_2008gaussian} where its mean and
variance are given by 
\begin{align*}
\mu\left(\bx_{*}\right)= & \mathbf{k}_{*}\mathbf{K}^{-1}\mathbf{y}\\
\sigma^{2}\left(\bx_{*}\right)= & k_{**}-\mathbf{k}_{*}\mathbf{K}^{-1}\mathbf{k}_{*}^{T}
\end{align*}
GPs provide a full probabilistic model of the data, and allow us to
compute not only the model\textquoteright s prediction at input points
but also to quantify the uncertainty in the predictions. Therefore,
GP is flexible as a nonparametric prior for Bayesian optimization.

\subsection{Bayesian optimization}

\begin{figure}
\begin{centering}
\includegraphics[width=0.7\columnwidth]{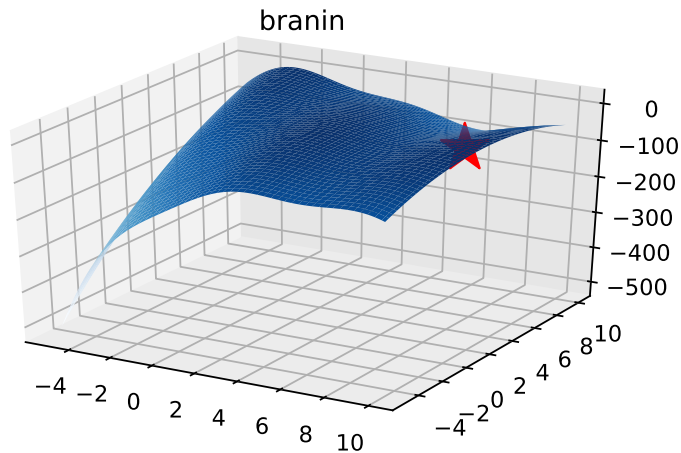}
\par\end{centering}
\caption{Example of the Branin function. The global maximum is denoted as the
red star. \label{fig:Example_Branin}}

\end{figure}

We assume that $f$ is a black-box function, that is, its form is
unknown and further it is expensive to evaluate. Perturbed evaluations
of the type $y_{i}=f\left(\bx_{i}\right)+\epsilon_{i}$, where $\epsilon_{i}\sim\mathcal{N}\left(0,\sigma^{2}\right)$,
are available. Example of the Branin function is in Fig. \ref{fig:Example_Branin}.
Bayesian optimization makes a series of evaluations $\bx_{1},...,\bx_{T}$
of $f$ such that the maximum of $f$ is found in the fewest iterations
\cite{Snoek_2012Practical,Shahriari_2016Taking,Dai_2016Cascade,Nguyen_ICDM2016Budgeted}.
Formally, let $f:\mathcal{X}\rightarrow\mathcal{R}$ be a well behaved
function defined on a compact subset $\mathcal{X}\subseteq\mathcal{R}^{D}$.
Our goal is solving the global optimization problem
\begin{align}
\bx^{*} & =\underset{\bx\in\mathcal{X}}{\text{argmax}\thinspace f\left(\bx\right)}.\label{eq:optimization_original}
\end{align}
Bayesian optimization reasons about $f$ by building a Gaussian process
through evaluations \cite{Rasmussen_2006gaussian}. This flexible
distribution allows us to associate a normally distributed random
variable at every point in the continuous input space. 

\subsubsection*{Acquisition functions}

As the original function is expensive to evaluate, the acquisition
function acts as a surrogate that determines which point should be
selected next. Therefore, instead of maximizing the original function
$f$, we maximize the acquisition function $\alpha$ to select the
next point to evaluate
\begin{align*}
\bx_{t+1} & =\argmax{x\in\mathcal{X}}\alpha_{t}\left(\bx\right).
\end{align*}
In this auxiliary maximization problem, the objective is known and
easy to evaluate and can be easily carried out with standard numerical
techniques such as multi-start or DIRECT \cite{Jones_1993Lipschitzian}.
We consider the case of computing the acquisition function $\alpha\left(\bx\right)$
from the posterior distribution of GP. These acquisition functions
are carefully designed to trade off exploration of the search space
and exploitation of current promising regions.  Among many existing
acquisition functions in literature \cite{Hennig_2012Entropy,Hernandez_2014Predictive,Srinivas_2010Gaussian,Mockus_1978Application,Jones_2001Taxonomy,Freitas_2012Exponential,Nguyen_2016Think},
we briefly describe three common acquisition functions including probability
of improvement, expected improvement and upper confidence bound. The
early work of \cite{Kushner_1964New} suggested maximizing the probability
of improvement (PI) over the incumbent $\alpha^{\textrm{PI}}(\bx)=\Phi\left(\frac{\mu\left(\bx\right)-y^{+}}{\sigma\left(\bx\right)}\right)$,
where $\Phi$ is the standard normal cumulative distribution function
(cdf) and the incumbent $y^{+}=\max_{x_{i}\in\mathcal{D}_{t}}f\left(x_{i}\right)$.
However, the PI exploits quite aggressively. Thus, one could instead
measure the expected improvement (EI) \cite{Mockus_1978Application,Jones_1998Efficient}.
The expected improvement chooses the next point with highest expected
improvement over the current best outcome, i.e. the maximizer of $\alpha^{\textrm{EI}}\left(\bx\right)=\left(\mu\left(\bx\right)-\tau\right)\Phi\left(u\right)+\sigma\left(\bx\right)\phi\left(u\right)$
where $\phi$ is the standard normal probability distribution function
(pdf), $\tau$ is the current best value, $u\left(x\right)=\frac{\mu\left(\bx\right)-\tau}{\sigma\left(\bx\right)}$,
for $\sigma>0$ and zero otherwise. The Gaussian process upper confidence
bound (GP-UCB) acquisition function is given by $\alpha^{\textrm{GP-UCB}}(\bx)=\mu(x)+\sqrt{\beta}\sigma(x)$,
where $\sqrt{\beta}$ is a domain-specific positive parameter which
trades exploration and exploitation. There are theoretically motivated
guidelines for setting and scheduling the hyper-parameter $\sqrt{\beta}$
to achieve sublinear regret \cite{Srinivas_2010Gaussian}.

\begin{figure}
\begin{centering}
\includegraphics[width=0.95\columnwidth]{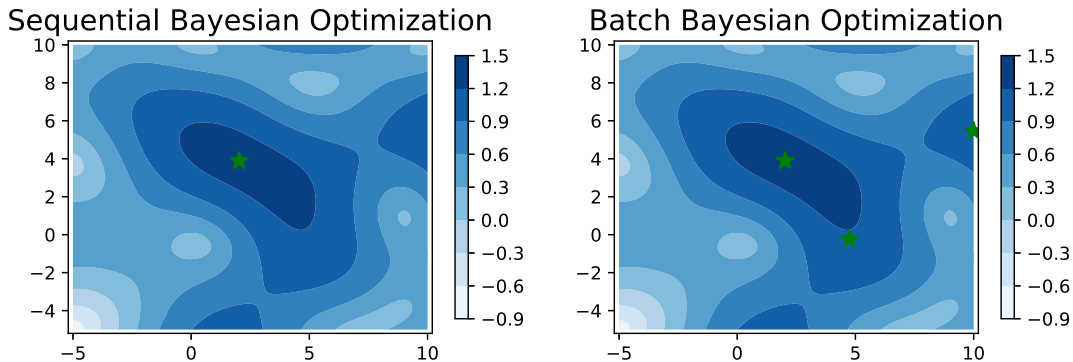}
\par\end{centering}
\caption{Examples of sequential Bayesian optimization selecting one location
versus batch Bayesian optimization selecting multiple (three) locations
at each iteration. The selected points are indicated by the green
stars. The curve is the acquisition function using UCB on Branin function.\label{fig:Examples_sequential_batch}}

\end{figure}

\subsection{Batch Bayesian optimization}

Bayesian optimization is conventionally posed as a sequential problem
where each experiment is completed before taking a new one. In practice
it may be advantageous to run multiple function evaluations in parallel.
Therefore, we consider evaluating $f$ using a batch of points. An
example of batch Bayesian optimization versus sequential Bayesian
optimization is illustrated in Fig. \ref{fig:Examples_sequential_batch}.
As discussed earlier, such scenarios appear, for instance, in the
optimization of computer models where several machines (or cores)
are available to run experiments in parallel, or in wet-lab experiments
when the time of testing one experimental design is the same as testing
a batch. Formally, we maximize the acquisition function by finding
a collection of points as
\begin{align}
\bX_{t} & =\left[\bx_{t1},\bx_{t2},...,\bx_{tn_{t}}\right]=\argmax{\bx\in\mathcal{X}}\alpha_{t-1}\left(\bx\right)\label{eq:maximizing_acquisition_batch}
\end{align}
where $n_{t}$ is the batch size at iteration $t$. Existing approaches
often fix the batch size $n_{t}$ to a constant value for all iterations.

\subsection{Existing batch Bayesian optimization methods}

Since the optimization in Eq. \ref{eq:maximizing_acquisition_batch}
is intractable, batch BO techniques avoid this computational burden
by resorting to different strategies. The first BO parallelization
was used in the context of the learning Bayesian networks \cite{Ovcenavsek_2000Parallel},
to the best of our knowledge. Then batch BO simulation matching \cite{Azimi_2010Batch}
aims to select a batch of size $q$, which includes points \textquoteleft close
to\textquoteright{} the best point. 

Developing from the expected improvement acquisition function (EI),
the multi-point expected improvement (q-EI)\cite{Ginsbourger_2007Multi}
treats the acquisition function as the conditional expectation of
the improvement obtained by $q$ points. Constant Liar (CL) is a heuristic
q-EI algorithm \cite{Ginsbourger_2008Multi}, which uses a greedy
approach to iteratively construct a batch of $q$ points. At each
iteration, the heuristic uses the sequential algorithm to find a point
that maximizes the expected improvement as follows: first, the maximum
of the acquisition is found and to move to the next maximum by suppressing
this point. This is done by inserting the outcome at this point as
a constant value. The GP posterior is updated through the ``fake''
outcome which then is used to compute the acquisition function. This
process is repeated until the batch is filled. Further parallel extension
for q-EI are proposed in \cite{Frazier_2012Parallel,Wang_2015Parallel}.

Another direction \cite{Contal_2013Parallel,Desautels_2014Parallelizing}
in batch Bayesian optimization exploits an interesting fact about
GPs: the predictive variance of GPs depends only on the feature $\bx$,
but not the outcome values $\by$. The GP-BUCB algorithm \cite{Desautels_2014Parallelizing}
and GP-UCB-PE \cite{Contal_2013Parallel} extend the sequential UCB
to a batch setting by first selecting the next point, updating the
predictive variance which in turn alters the acquisition function,
and then selecting the next point. This is repeated till the batch
is filled. In particular, the GP-UCB-PE \cite{Contal_2013Parallel}
chooses the first point of the batch via the UCB score and then defines
a \textquotedblleft relevance region\textquotedblright{} and selects
the remaining points from this region greedily to maximize the information
gain, to focus on pure exploration (PE).

Batch BO can also be developed using information-based policies \cite{Hernandez_2014Predictive}.
PES aim to select the point which maximizes the information gain of
the model. This is solved by expressing the expected reduction in
the differential entropy of the predicted distribution. Parallel Predictive
Entropy Search (PPES) \cite{Shah_2015Parallel} extends the Predictive
Entropy Search (PES) algorithm of \cite{Hernandez_2014Predictive}
to a batch setting. 

More recently, Local Penalization (LP) \cite{Gonzalez_2015Batch}
presents a heuristic approach for batch BO by iteratively penalizing
the current peak in the acquisition function to find the next peak.
LP depends on the estimation of Lipschitz constant to flexibly penalize
the peaks. However, in general scenarios, the Lipschitz constant $L$
is unknown. For ease of computation, \cite{Gonzalez_2015Batch} estimates
the Lipschitz constant of the GP predictive mean instead of the actual
acquisition function. Furthermore, the use of a unique value of $L$
assumes that the function is Lipschitz homoscedastic. Gonzalez et
al \cite{Gonzalez_2015Batch} has demonstrated that LP outperforms
a wide range of baselines in batch Bayesian optimization. We thus
consider LP as the most competitive baseline in batch BO.

Although the batch algorithms can speedup the selection of experiments,
there are two main drawbacks. First, most of the proposed batch BO
\cite{Ginsbourger_2008Multi,Azimi_2010Batch,Azimi_2012Hybrid,Contal_2013Parallel,Desautels_2014Parallelizing,Gonzalez_2015Batch}
involve a greedy algorithm, which chooses individual points until
the batch is filled. This is often detrimental as the requirement
to fill a batch enforces the selection of not only the true maxima
but also noisy peaks. Second, most approaches use a predefined and
fixed batch size $q$ for all iterations. Fixed batch size may be
inefficient because the number of peaks in the acquisition function
is unknown and constantly changing when the GP posterior is updated.
For example, if a function has two real maxima, a batch size of more
than two will force the selection of noisy points. Therefore, time
and resources are wasted for evaluating these noisy points that contravening
the goal of BO is to save the number of evaluations. In contrast,
if we under-specify the number of peaks, we could miss important points
that may be the optimal solution of $f$. Hence, it is necessary to
identify the appropriate, if not exact, number of peaks for evaluation
while preserving the ability of finding the optimal value of $f$.

\section{Budgeted Batch Bayesian Optimization}

We propose the novel batch Bayesian optimization method that learns
the suitable batch size for each iteration. We term our approach \emph{budgeted
batch Bayesian optimization (B3O)}. The proposed method is economic
in terms of number of optimal evaluations whilst preserving the performance. 

We first describe our approach to approximate the acquisition function
using the infinite Gaussian mixture model (IGMM). Then, we present
the batch generalized slice sampler to efficiently draw samples under
the acquisition function. Next, we briefly describe the existing variational
inference technique for IGMM. Finally, we summarize our algorithm.

\subsection{Batch Bayesian optimization}

We consider the batch Bayesian optimization. As a Bayesian optimization
task, our ultimate goal is solving the global optimization problem
of finding $\bx^{*}=\underset{\bx\in\mathcal{X}}{\text{argmax}\thinspace f\left(\bx\right)}$
by making a series of batch evaluations $\bX_{t},\bX_{2}...\bX_{T}$
where $\bX_{t}=\left[\bx_{t1},\bx_{t2},...,\bx_{tn_{t}}\right]$ such
that the maximum of $f$ is found \cite{Snoek_2012Practical,Shahriari_2016Taking}
where $n_{t}$ is the batch size. 

\subsection{Approximating the acquisition function with infinite Gaussian mixture
models}

\begin{figure*}
\begin{centering}
\includegraphics[width=0.92\columnwidth]{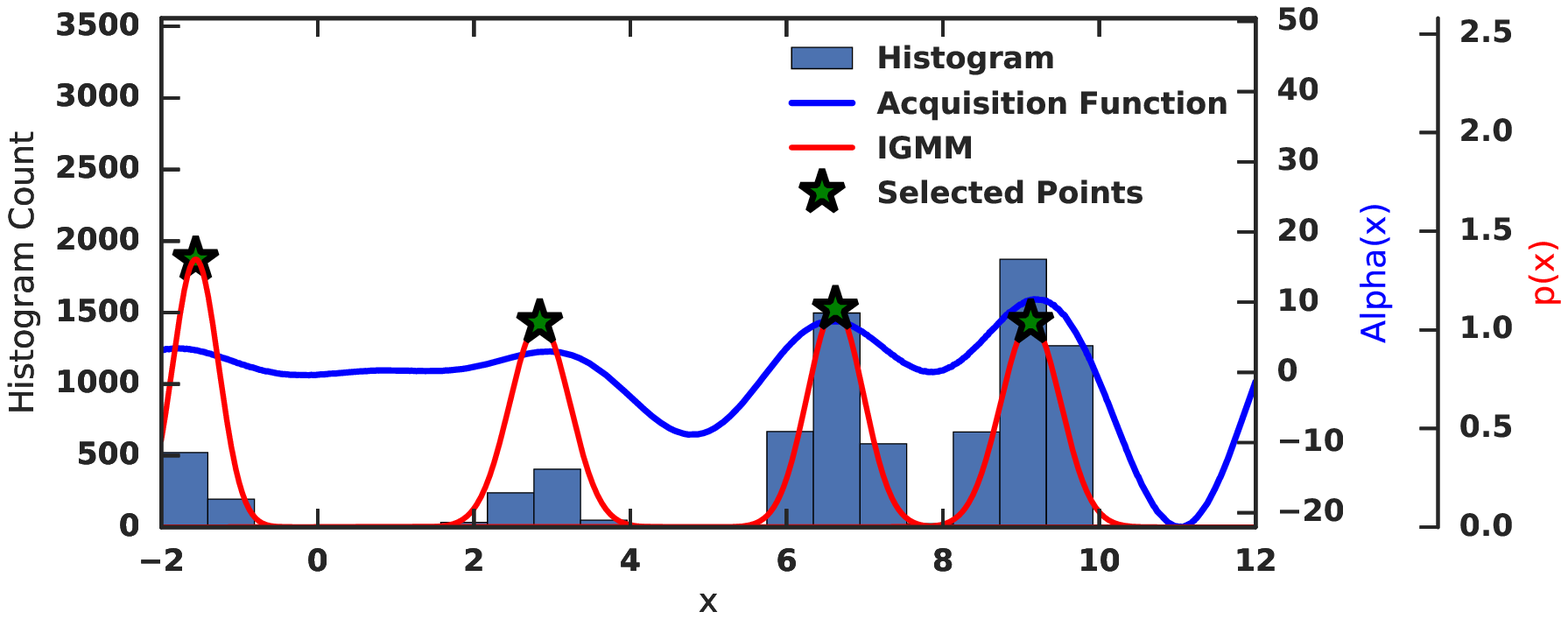}
\par\end{centering}
\centering{}\includegraphics[width=0.92\columnwidth]{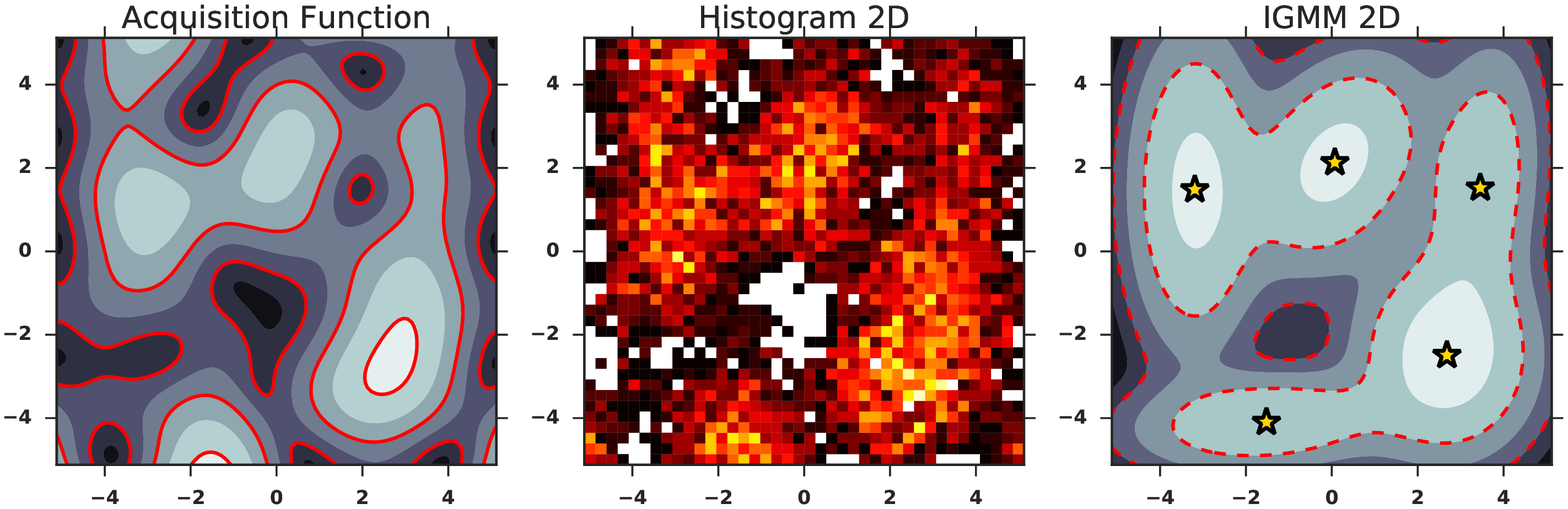}\caption{Budgeted Batch Bayesian Optimization. Top: i.i.d samples are drawn
(see the histogram) from the acquisition function $\alpha$ (\textcolor{blue}{blue}
line) using the proposed batch generalized slice sampling. These generated
samples after fitted with IGMM (\textcolor{red}{red} line) and estimated
means are selected (\textcolor{green}{green} star). Bottom: Illustration
in 2D of B3O.\label{fig:B3O_Illustration}}
\end{figure*}

The acquisition function $\alpha$ is often multi-modal with the unknown
number of peaks. It is intuitive to assume that the acquisition function
can be approximated by a mixture of Gaussians (cf. Fig. \ref{fig:B3O_Illustration})
wherein the peaks in the acquisition function are equivalent to the
mean locations of the underlying Gaussians. Because the number of
peaks are unknown, we borrow the elegance of Bayesian nonparametrics
\cite{Hjort_etal_bk10_bayesian} to identify the unknown number of
Gaussian components. We use the infinite Gaussian mixture model (IGMM)
\cite{Rasmussen_1999_Infinite} which induces the Dirichlet Process
prior over possibly infinite number of Gaussian components.  In
IGMM, each Gaussian component representing a peak $k$ is parameterized
by the mean $\mu_{k}$ and the covariance $\Sigma_{k}$. It is interesting
that the estimated mean $\mu_{k}$ is the location of the peaks while
the estimated covariance $\Sigma_{k}$ will capture the shape of the
peaks in the acquisition function. In contrast, previous work \cite{Gonzalez_2015Batch}
relies on an unique Lipschitz constant to represent the shape of the
peak that is not reasonable for the heteroscedastic setting when the
peaks have different shapes. Formally, the probability density function
of the IGMM is defined as follows $\sum_{k=1}^{\infty}\pi_{k}\times\mathcal{N}\left(s\mid\mu_{k},\Sigma_{k}\right)$
where $s$ is an observation, $\pi_{k}$ is the mixing proportion
in IGMM and $\mathcal{N}\left(s\mid\mu_{k},\Sigma_{k}\right)$ is
the Gaussian density function of the component $k$.

Our primary goal in utilizing IGMM is to approximately find the mean
$\mu_{k}$(s) of Gaussian distributions as the unknown peaks from
the acquisition function. However, directly estimating IGMM from acquisition
function is intractable. Therefore, we use the intermediate step to
draw samples under the acquisition function and then fit the IGMM
to learn the mean locations as the batch of points to evaluate.

\subsection{Batch generalized slice sampler (BGSS) \label{subsec:Batch-Slice-Sampler}}

\begin{algorithm}[h]
\begin{algor}
\item [{{*}}] Input: \#MaxIter, acquisition function $\alpha$, $\bx_{\textrm{min}},\bx_{\textrm{max}}$
\end{algor}
\begin{algor}[1]
\item [{{*}}] $\bs_{0}\sim\textrm{uniform}\left(\bx_{\textrm{min}},\bx_{\textrm{max}}\right)$ 
\item [{{*}}] $\alpha_{\textrm{min}}=\min_{x\in\mathcal{X}}\alpha\left(x\right)$
\item [{for}] $i=1$ to $\textrm{\#MaxIter}$
\item [{{*}}] $u_{i}\sim\textrm{uniform}\left(\alpha_{\textrm{min}},\alpha\left(\bs_{i-1}\right)\right)$
\item [{while}] (notAccept)
\item [{{*}}] $\bs_{i}\sim\textrm{uniform}\left(\bx_{\textrm{min}},\bx_{\textrm{max}}\right)$ 
\item [{if}] $\alpha(\bs_{i})>u_{i}$ \#Accept
\item [{{*}}] $S=S\cup\bs_{i}$
\item [{endif}]~
\item [{endwhile}]~
\item [{endfor}]~
\end{algor}
\begin{algor}
\item [{{*}}] Output: $S$
\end{algor}
\caption{Algorithm for generalized slice sampling.\label{alg:Algorithm-for_BGSS}}
\end{algorithm}
We now present the sampling technique to draw samples under the acquisition
function for fitting to the IGMM. We note that the sampling process,
by its nature, generates more samples from high probability regions.
This implies that most of the samples come from the peaks. Thus, even
with small number of samples, the sampling process approximates the
peaks and is then efficient. There are many existing sampling methods
to draw samples from the probability distributions, for example, Metropolis-Hastings
\cite{Metropolis_1953equation}. However, to sample under the $D$-dimensional
acquisition function which is not strictly probability distribution,
we utilize the accept-reject sampling \cite{Casella_2004Generalized}
where we keep samples in the region under the density function and
ignore samples if they are outside the curve.

\begin{figure}
\begin{centering}
\includegraphics[width=0.7\columnwidth]{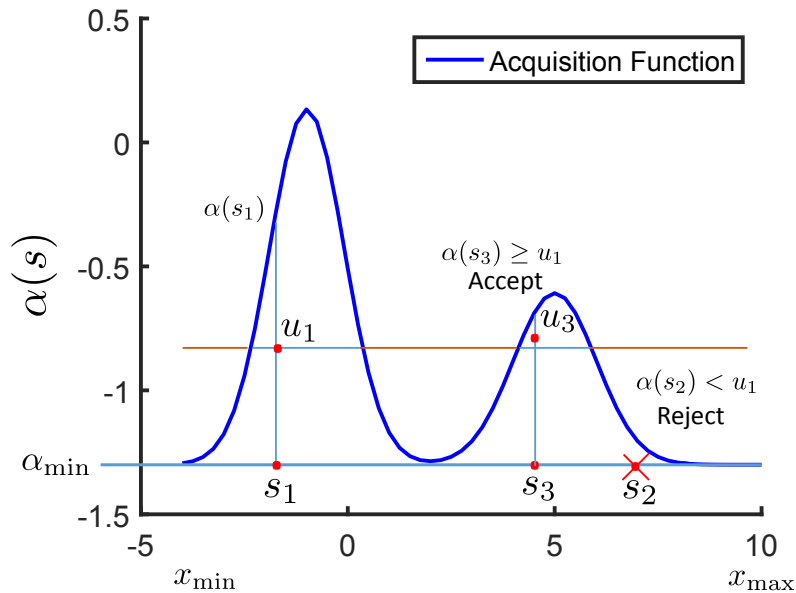}
\par\end{centering}
\caption{The illustration of the proposed generalized slice sampling to draw
samples under the acquisition functions. We consider an example as
follows. The first point $s_{1}$ is uniformly drawn from $\textrm{Uniform}\left(x_{\min},x_{\max}\right)$,
then $u_{1}\sim\textrm{Uniform}\left(\alpha_{\min},\alpha(s_{1})\right)$.
Next, the second point is sampled $s_{2}\sim\textrm{Uniform}\left(x_{\min},x_{\max}\right)$.
Due to $\alpha(s_{2})<u_{1}$, we reject $s_{2}$. We continue to
draw $s_{3}$ and accept it as $\alpha(s_{3})\ge u_{1}$. We repeat
the algorithm by again sampling $u_{3}\sim\textrm{Uniform}\left(\alpha_{\min},\alpha(s_{3})\right)$.
\label{fig:Illustration-of-GeneralizedSS}}
\end{figure}
In particular, we select to use the slice sampling \cite{Neal_03slice}
because it is easily implemented for univariate distributions, and
can be used to sample from a multivariate distribution by updating
each variable in turn.  Suppose we wish to sample from a distribution
for a variable, $x$, taking values in some subset of $\mathcal{R}^{D}$,
whose density is proportional to some functions $g(x)$. We can do
this by sampling uniformly from the $(D+1)$-dimensional region that
lies under the plot of $g(x)$. This idea can be formalized by introducing
an auxiliary real variable, $u$, and defining a joint distribution
over $x$ and $u$ that is uniform over the region $R=\left\{ (x,u):0<u<g(x)\right\} $
below the curve or surface defined by $g(x)$. The joint density for
$(x,u)$ is 
\begin{align*}
p\left(x,u\right) & =\begin{cases}
\frac{1}{Z} & \textrm{if}\thinspace0<u<g\left(x\right)\\
0 & \textrm{otherwise}
\end{cases}
\end{align*}
where $Z=\int g\left(x\right)dx$. The marginal density for $x$ is
then $p\left(x\right)=\int_{0}^{g(x)}\frac{1}{Z}du=\frac{g\left(x\right)}{Z}$.
To sample for $x$, we can sample jointly for $(x,u)$, and then ignore
$u$.

We extend the standard slice sampler to the \emph{generalized slice
sampler} that draws samples from $D$-dimensional acquisition function
$\alpha$ which is not a proper distribution and can be negative.
To draw $u$ uniformly over the region $R$ below curve of the acquisition
function, we get the $\alpha_{\textrm{min}}=\min_{x\in\mathcal{X}}\alpha\left(x\right)$
obtained by one of the non-convex optimization toolbox (e.g., DIRECT
\cite{Jones_1993Lipschitzian}). Therefore, we can define the joint
density
\begin{align*}
p\left(x,u\right)= & \begin{cases}
\frac{1}{Z} & \textrm{if}\thinspace\alpha_{\textrm{min}}<u<\alpha\left(x\right)\\
0 & \textrm{otherwise}
\end{cases}
\end{align*}
where $Z=\int\alpha\left(x\right)dx$. The marginal density for $x$
is then $p\left(x\right)=\int_{\alpha_{\textrm{min}}}^{\alpha(x)}\frac{1}{Z}du=\frac{\alpha\left(x\right)-\alpha_{\textrm{min}}}{Z}$.
To sample for $x$, we can sample jointly for $(x,u)$, and then ignore
$u$ following the Algorithm \ref{alg:Algorithm-for_BGSS} and Fig.
\ref{fig:Illustration-of-GeneralizedSS} where $\textrm{uniform}\left(\bx_{\textrm{min}},\bx_{\textrm{max}}\right)$
denotes for drawing randomly from the uniform distribution within
the bound of $\left[\bx_{\textrm{min}},\bx_{\textrm{max}}\right]$,
which is predefined.

Slice sampling approach \cite{Neal_03slice} also brings a challenge
for high-dimensional data that it is hard to find accepted area $R$
such that $g\left(R\right)>u$. There are two main reasons. First,
the set of accepted samples could be very small with respect to the
hyper-rectangle $R$. Second, $R$ usually lies in a high-dimensional
space, implying the curse of dimensionality. Therefore, it is a big
drawback of slice sampling to deal with high dimensions \cite{Tibbits_2011Parallel,Pietrabissa_2014Parallel}.

To overcome the problem in high-dimensional functions, we utilize
the generalized slice sampler in batch setting where a bunch of samples
are drawn i.i.d at different places. Since our goal is to draw a collection
of i.i.d. samples $x_{1,2,...N}$ from the acquisition function $\alpha$
(to later estimate the IGMM). We define a collection of joint density
distribution $p\left(x_{1},u_{1}\right),...p\left(x_{M},u\right)$
and perform the generalized slice sampling for each joint distribution
independently. In the experiment, we set $M$ as $200$. In Sec \ref{subsec:Analysis-of-BatchSliceSampling},
we further study the computational efficiency of the \emph{batch generalized
slice sampling} (BGSS) for drawing samples from high dimensional acquisition
functions (e.g., $D=50$).

\subsection{Variational inference for infinite Gaussian mixture model}

IGMM is the nonparametric mixture model where the prior distribution
over the mixing proportion is a Dirichlet process \cite{Ferguson_73bayesian}.
There are few existing approaches to learn a IGMM, such as collapsed
Gibbs sampler \cite{Kamper_2013Gibbs} and variational inference \cite{blei2006variational}.
In this paper, we follow \cite{blei2006variational} to derive the
variational inference for IGMM \cite{Rasmussen_1999_Infinite} since
the variational approach is generally faster than the Gibbs sampler.
After fitting the IGMM using variational inference, we obtain the
mean locations $\mu_{1....K}$ as the selected points for evaluations
in the batch BO setting. We note that $K$ is unknown and identified
automatically in this Bayesian nonparametric setting.

Using the collection of samples $\left(\bs_{i}\right)_{i=1}^{N}$
drawn from our batch generalized slice sampler, we consider the IGMM
generated as follows. We first draw a mixing proportion from the stick-breaking
process given the concentration parameter $\gamma$ as $\pi_{k}\sim v_{k}\prod_{l=1}^{k-1}\left(1-v_{l}\right)$
where $v_{k}\sim\betapdf\left(1,\gamma\right)$. Next, for each component
$k$ in the model, we sample the mean $\mu_{k}\sim\mathcal{N}\left(\mu_{0},\idenmat\right)$
and the covariance matrix $\Sigma_{k}\sim\wshpdf\left(\tau_{k},\idenmat\right)$.
Finally, we generate the assignment $z_{i}\sim\textrm{Stick}\left(\pi\right)$
and the observation $\bs_{i}\sim\mathcal{N}\left(\mu_{z_{i},}\Sigma_{z_{i}}\right)$. 

For posterior inference using variational inference, we first write
the bound on the likelihood of the data as follows:
\begin{align}
\log p\left(\bX\right)\ge & \ESS[\log p\left(v\right)]{}+\ESS[\log p\left(\mu\right)]{}+\ESS[\log p\left(\Sigma\right)]{}+\ESS[\log p\left(z\right)]{}+\ESS[\log p\left(x\right)]{}\nonumber \\
 & -\ESS[\log q\left(v,\mu,\Sigma,z\right)]{}\label{eq:LowerBound_VI_IGMM}
\end{align}
To define the variational distribution $q$ in the Eq. (\ref{eq:LowerBound_VI_IGMM}),
we need to construct a distribution on an infinite set of random variables
$v_{k},$$\mu_{k}$ and $\Sigma_{k}$. For this approach be tractable,
we truncate the variational distribution at some value $K$ by setting
$q(v_{K}=1)=1$ and we can ignore $\mu_{k},\Sigma_{k}$ for $k>K$
\cite{blei2006variational}. 

Then, we factorize the variational distribution using mean field assumption
as the following $q\left(v,\mu,\Sigma,z\right)=\prod_{k=1}^{K}q\left(v_{k}\mid\eta_{k}\right)q\left(\mu_{k}\mid\lambda_{k}\right)q\left(\Sigma_{k}\mid a_{k},B_{k}\right)\prod_{i=1}^{N}q\left(z_{i}\mid\phi_{i}\right)$.
Particularly, the variational distributions for these variables are
defined as $q\left(v_{k}\mid\eta_{k}\right)=\betapdf\left(\eta_{k1},\eta_{k2}\right),q\left(\mu_{k}\mid\lambda_{k}\right)=\mathcal{N}\left(\lambda_{k},\idenmat\right)$,
$q\left(\Sigma_{k}\right)=\textrm{Wishart}(a_{k},B_{k})$ and $q\left(z_{i}\mid\phi_{i}\right)=\multpdf\left(\phi_{i}\right)$.

 After factorizing the variational distribution using mean field,
we maximize the lower bound in Eq. (\ref{eq:LowerBound_VI_IGMM}).
We first take the partial derivative w.r.t. each variational variables.
We then equate it to zero and solve the optimization for each variables.
Due to the space restriction, we shall refer the detailed inference
of the infinite Gaussian mixture model to \cite{Rasmussen_1999_Infinite,blei2006variational}.

\subsection{Algorithm}

\begin{algorithm}
\begin{algor}
\item [{{*}}] Input: $\mathcal{D}_{0}=\left\{ \bx_{i},y_{i}\right\} _{i=1}^{n_{0}}$,
\#iter $T$, acquisition function $\alpha$
\end{algor}
\begin{algor}[1]
\item [{for}] $t=1$ to $T$
\item [{{*}}] Fit a GP from the data $\mathcal{D}_{t}$.
\item [{{*}}] Build the acquisition function $\alpha\left(\right)$ from
GP.
\item [{{*}}] Draw auxiliary samples $\bs\sim\alpha\left(\bx\right)$ from
Algorithm \ref{alg:Algorithm-for_BGSS}.
\item [{{*}}] Fit the Infinite GMM using $\bs$.
\item [{{*}}] Obtain batch $\bX_{t}\leftarrow[\mu_{1},...,\mu_{n_{t}}]$
where $\mu_{k}$ is the estimated mean atom from IGMM.
\item [{{*}}] $\bY_{t}=[y_{t,1},...y_{t,n_{t}}]\leftarrow$parallel evaluations
of $f\left(\bX_{t}\right)$.
\item [{{*}}] $\mathcal{D}_{t+1}=\mathcal{D}_{t}\cup\left[\left(\bx_{t,j},y_{t,j}\right)\right]_{j=1}^{n_{t}}$
\item [{endfor}]~
\end{algor}
\begin{algor}
\item [{{*}}] Output: $\mathcal{D}_{T}$
\end{algor}
\caption{Algorithm for Budgeted Batch Bayesian Optimization (B3O)\label{alg:Algorithm-for-B3O}.}
\end{algorithm}
Although the technique of variational inference for IGMM and the slice
sampling are existed for years, the idea of connecting these things
at the right place for batch BO is novel. We summarize the steps
for B3O in Algorithm \ref{alg:Algorithm-for-B3O}. At an iteration
$t$, we will find a batch including $n_{t}$ points where the number
of point $n_{t}$ varies at each iteration.

We note that steps 2,3,7 and 8 are standard in Bayesian optimization
techniques. Our proposed method is highlighted in steps 4,5 and 6
to find a batch of points $\bX_{t}$ from the acquisition function.
Particularly, Fig. \ref{fig:B3O_Illustration} illustrates and summarizes
our steps 4,5 and 6 in 1D and 2D, respectively. We summarize the
computation time of steps 4, 5, and 6 in Sec \ref{subsec:ComparisonComputationalTime}
and further analyze step 4 by simulation in Sec \ref{subsec:Analysis-of-BatchSliceSampling}
which requires more computation than other steps.

\section{Experiments}

We evaluate the B3O algorithm using both synthetic functions and real-world
experimental settings. For the synthetic function evaluation, we utilize
8 functions across dimensions (1-10). In addition, we perform hyper-parameter
tuning for three machine learning algorithms - support vector regression
\cite{Smola_1997Support}, Bayesian nonparametric multi-label classification
\cite{Nguyen_2016Bayesian} and multi layer perceptron \cite{Ruck_1990Multilayer}.
We further consider real-world experimental design for Aluminum-Scandium
hardening. This heat treatment design involves searching for the best
combination of temperatures and times in two stages to achieve the
requisite alloy properties - in our case, hardness. 

We compare our results with baselines on best-found-value and the
total number of evaluations, given a fixed number of iterations. We
show that we outperform baselines in the best-found-value whilst requiring
fewer evaluations. We compare the performance of our proposed approach
with the state-of-the-art methods for batch BO.

\subsubsection*{Baselines}
\begin{itemize}
\item \emph{Expected Improvement (EI)\cite{Mockus_1978Application,Jones_2001Taxonomy}:
}this is a sequential approach using EI for the acquisition function.
\item \emph{GP-Upper Confident Bound (UCB)\cite{Srinivas_2010Gaussian}:}
this is a sequential approach using UCB with $\sqrt{\beta}=2$.
\item \emph{Gaussian process - Batch Upper Confidence Bound (GP-BUCB)} \cite{Desautels_2014Parallelizing}:
this is a batch BO utilizing the variance of GP for finding the next
point until the batch is filled.
\item \emph{Rand (EI and UCB): }this optimizes the acquisition function
to find the first element in the batch (this step is identical to
the sequential BO), then chooses a random sample within the bounds
until the batch is filled.
\item \emph{Constant Liar (EI and UCB) \cite{Ginsbourger_2008Multi,Ginsbourger_2010Kriging}:}
CL uses the predictive mean (from GP) to obtain new batch elements,
implemented in GPyOpt toolbox.
\item \emph{Local Penalization (LP) }\cite{Gonzalez_2015Batch}: This is
currently the state-of-the-art method for batch BO which has been
demonstrated to outperform most of other baselines \cite{Gonzalez_2015Batch}.
The source code is available in GPyOpt toolbox\footnote{https://github.com/SheffieldML/GPyOpt}.
\end{itemize}

\subsubsection*{Experimental Settings}

We denote the number of iterations by $T$, and the number of evaluations
by $N$. For the sequential setting of BO, the number of iterations
is identical to the number of evaluations, that is $N=T$ while $N>T$
for batch setting. The number of initial points $n_{0}$ for all methods
is set as $3\times D$ where $D$ is the dimension. For fixed-batch
approaches, the batch size at each iteration is set as $n_{t}=3$
for functions with $D<5$ or $n_{t}=D$ for functions with $D\ge5$.
In contrast, our method automatically determines the batch size $n_{t}$
which could differ at each iteration. Thus, we do not need to set
the batch size $n_{t}$ for B3O in advance. The performance of the
algorithms is compared for a fixed number of iterations $T$, given
as $10\times D$, and the total number of evaluated points is $N=\sum_{t=0}^{T}n_{t}.$

\begin{table}
\begin{centering}
\begin{tabular}{|c|c|c|}
\hline 
\textbf{Function, Dimension} & \textbf{Objective function $f\left(x\right)$} & \textbf{Optimum} $f\left(x^{*}\right)$\tabularnewline
\hline 
\hline 
Forrester, $D=1$ & $f\left(x\right)=\left(6x-2\right)^{2}\sin\left(12x-4\right)$ & $-6$\tabularnewline
\hline 
Dropwave, $D=2$ & $f\left(x\right)=-\frac{1+\cos\left(12\sqrt{x_{1}^{2}+x_{2}^{2}}\right)}{0.5\left(x_{1}^{2}+x_{2}^{2}\right)+2}$ & $-1$\tabularnewline
\hline 
\multirow{3}{*}{Hartmann, $D=3,6$} & $-\sum_{i=1}^{4}\alpha_{i}\exp\left(-\sum_{j=1}^{6}A_{ij}(x_{j}-P_{ij})^{2}\right)$ & $-3.86276$ ($D=3$)\tabularnewline
 & $\alpha=\left(1.0,1.2,3.0,3.2\right)^{T}$ & $-3.32237$ ($D=6$)\tabularnewline
 & {\small{}} & \tabularnewline
\hline 
Alpine2, $D=5,10$ & $\prod_{i=1}^{D}\sin\left(x_{i}\right).\sqrt{x_{i}}$ & -$2.808^{d}$\tabularnewline
\hline 
\multirow{2}{*}{gSobol, $D=5,10$} & $f(x)=\prod_{i=1}^{d}\frac{\left|4x_{i}-2\right|+a_{i}}{1+a_{i}}$ & $0$\tabularnewline
 & $a_{i}=1,\forall i=1...d$ & \tabularnewline
\hline 
\end{tabular}
\par\end{centering}
\centering{}\caption{Benchmark functions and dimensions used ranging from $1$ to $10$.\label{tab:Experimental-functions}}
\end{table}
In all the experiments, the squared exponential (SE) kernel given
as $k(x,x')=\exp\left(-\gamma||x-x'||^{2}\right)$ is used in the
underlying GP, where $\gamma=0.1\times D$ and $D$ is the dimension.
For methods using the UCB, $\sqrt{\beta}$ is fixed to 2 (following
the setting used in \cite{Gonzalez_2015Batch}), which allows us to
compare the different batch designs using the same acquisition function.
The results are taken over 20 replicates with different initial values.
We always maximize the objective function, maximizing $-f$ for cases
in which the goal is to find the minimum. All implementations are
in Python. All simulations are done on Windows machine Core i7 Ram
24GB.
\begin{figure*}
\begin{centering}
\includegraphics[width=0.51\columnwidth]{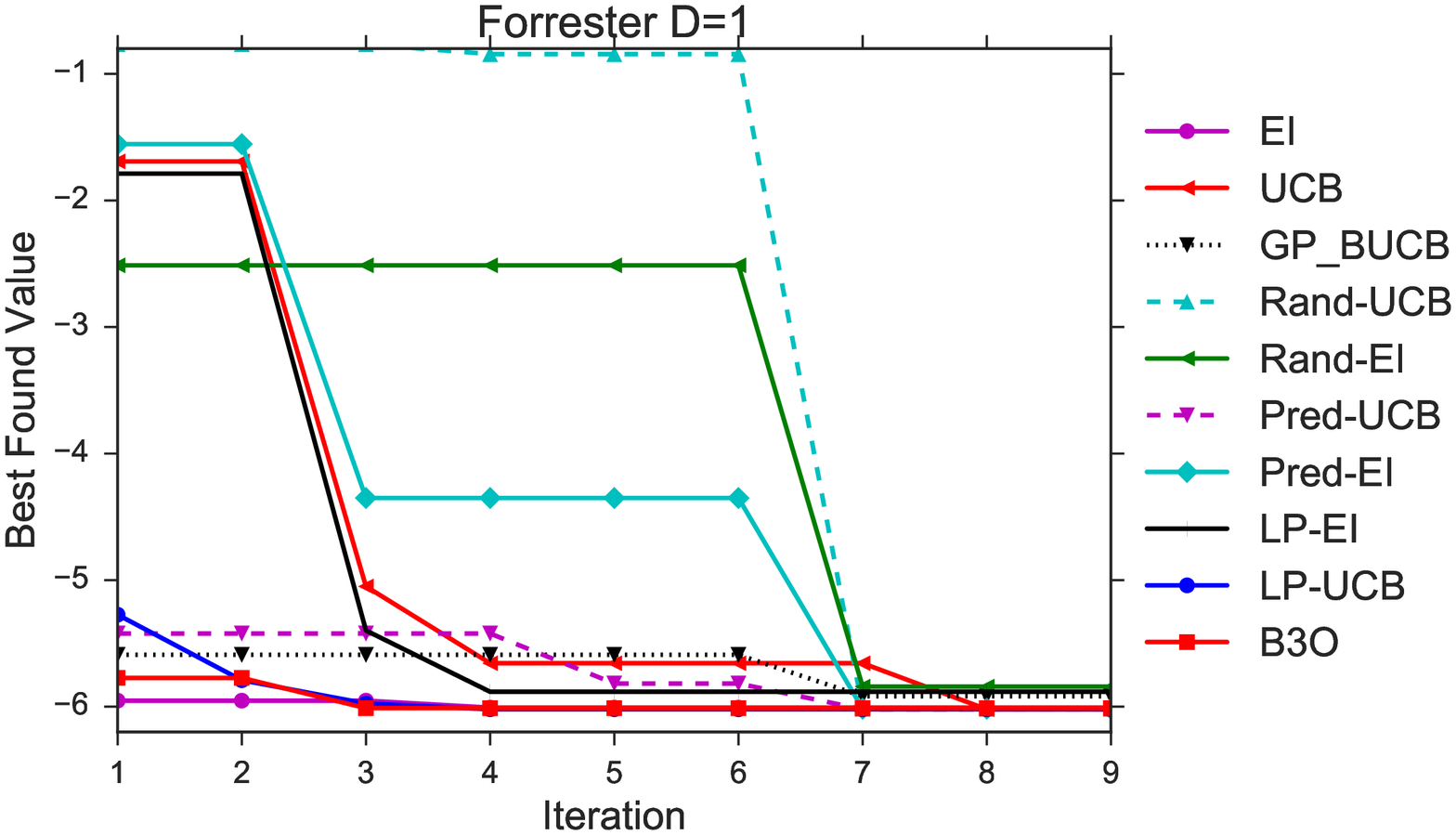}\hfill{}\includegraphics[width=0.48\columnwidth]{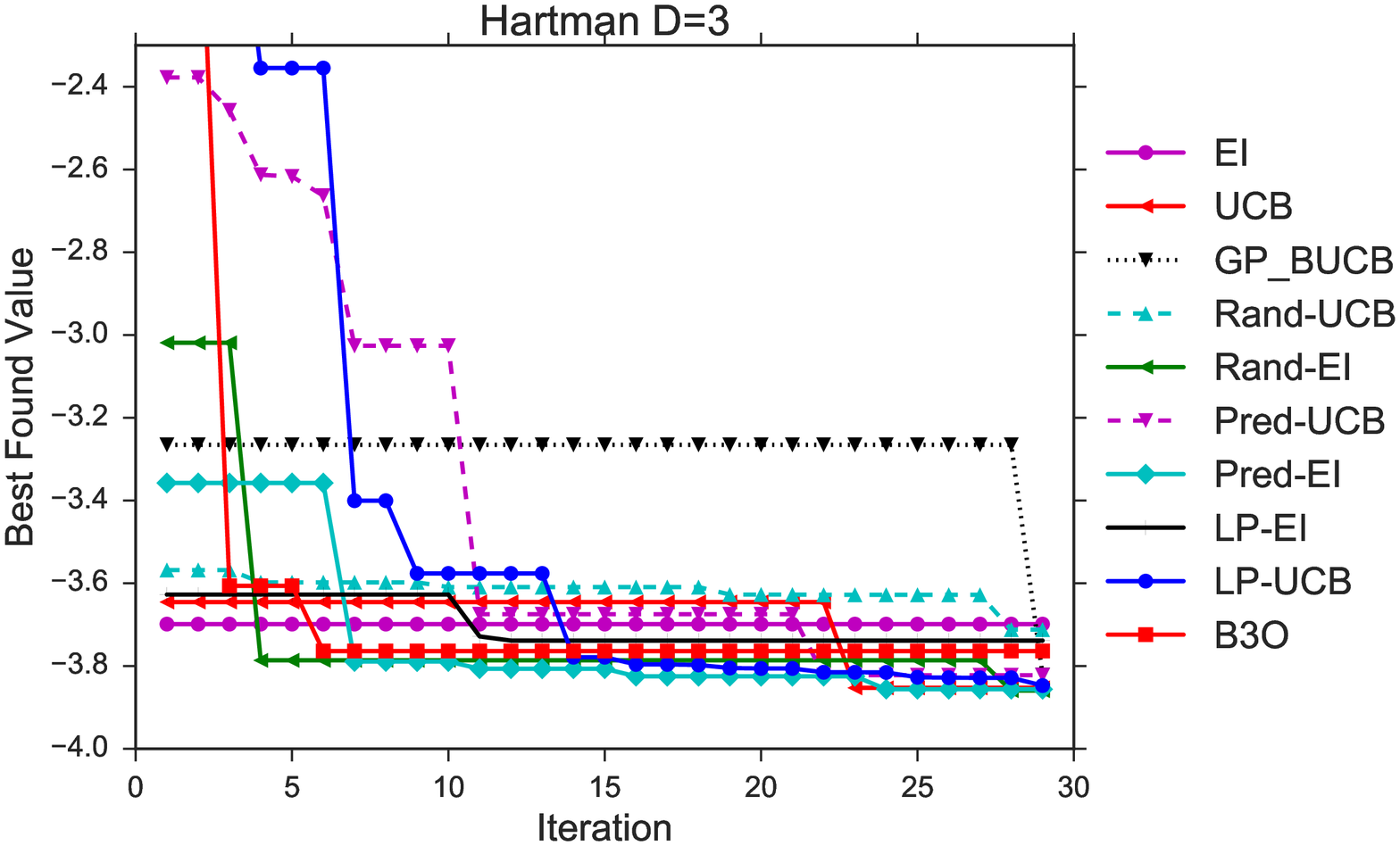}
\par\end{centering}
\begin{centering}
\includegraphics[width=0.51\columnwidth]{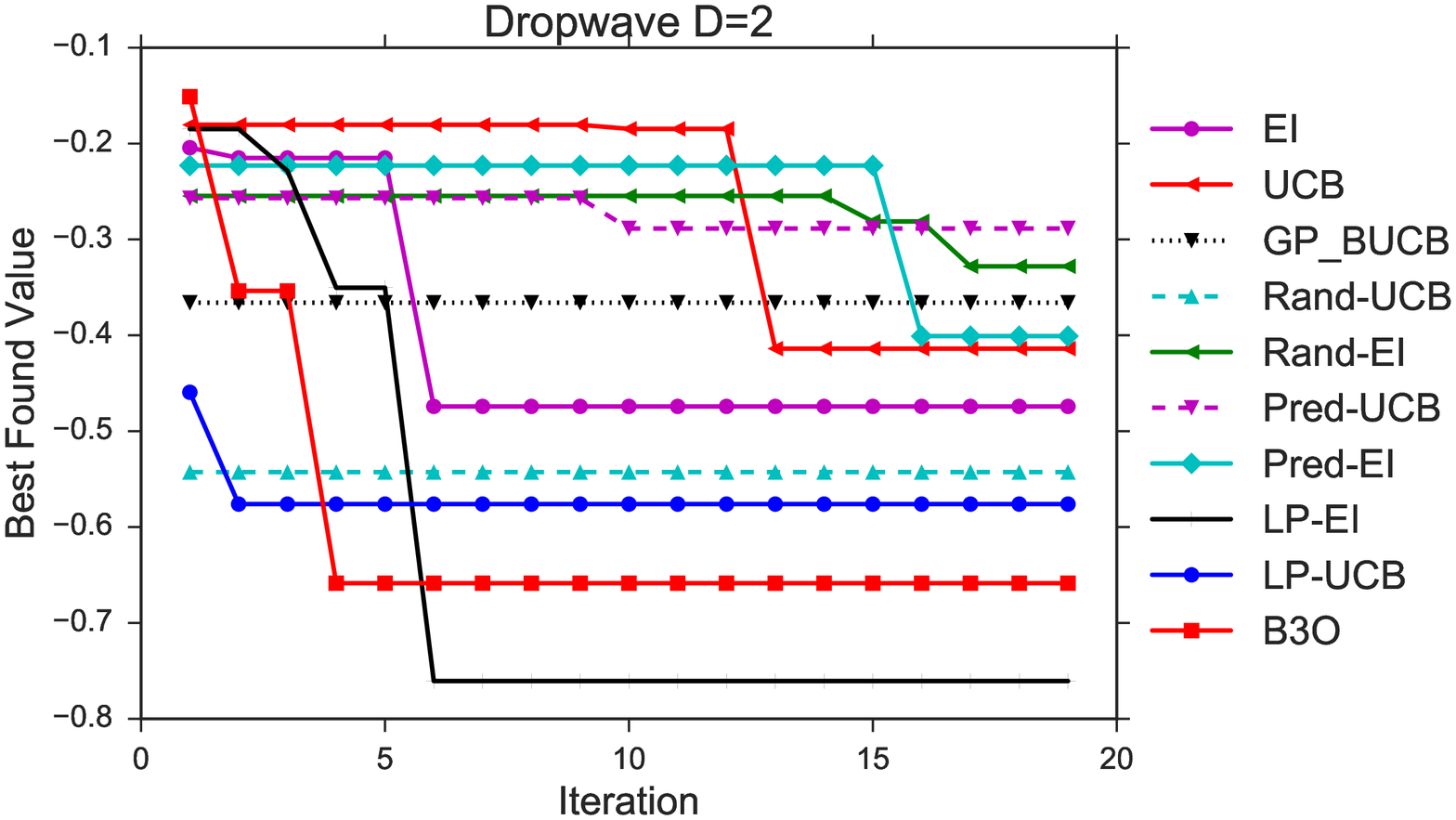}\hfill{}\includegraphics[width=0.48\columnwidth]{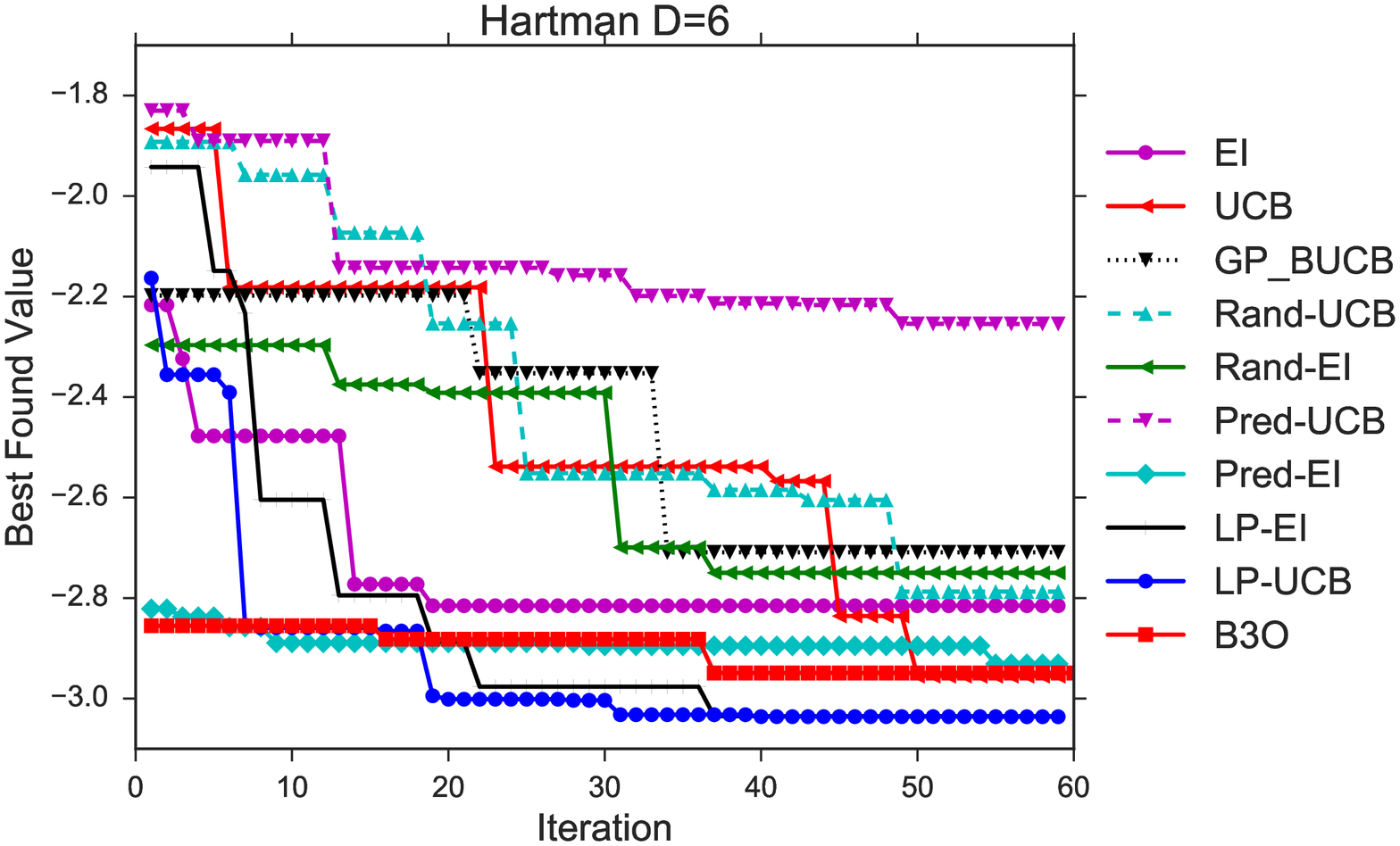}
\par\end{centering}
\begin{centering}
\includegraphics[width=0.51\columnwidth]{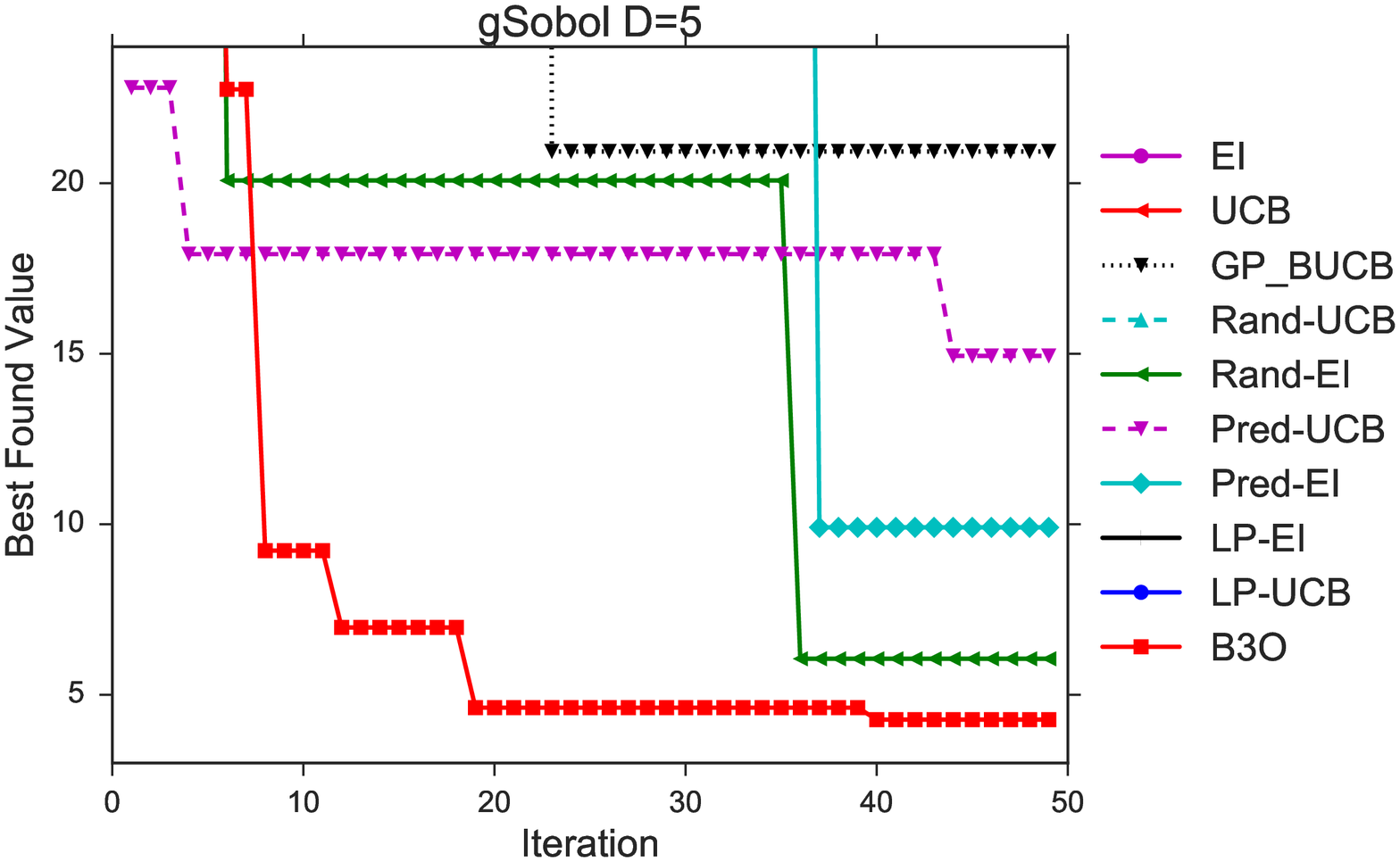}\hfill{}\includegraphics[width=0.48\columnwidth]{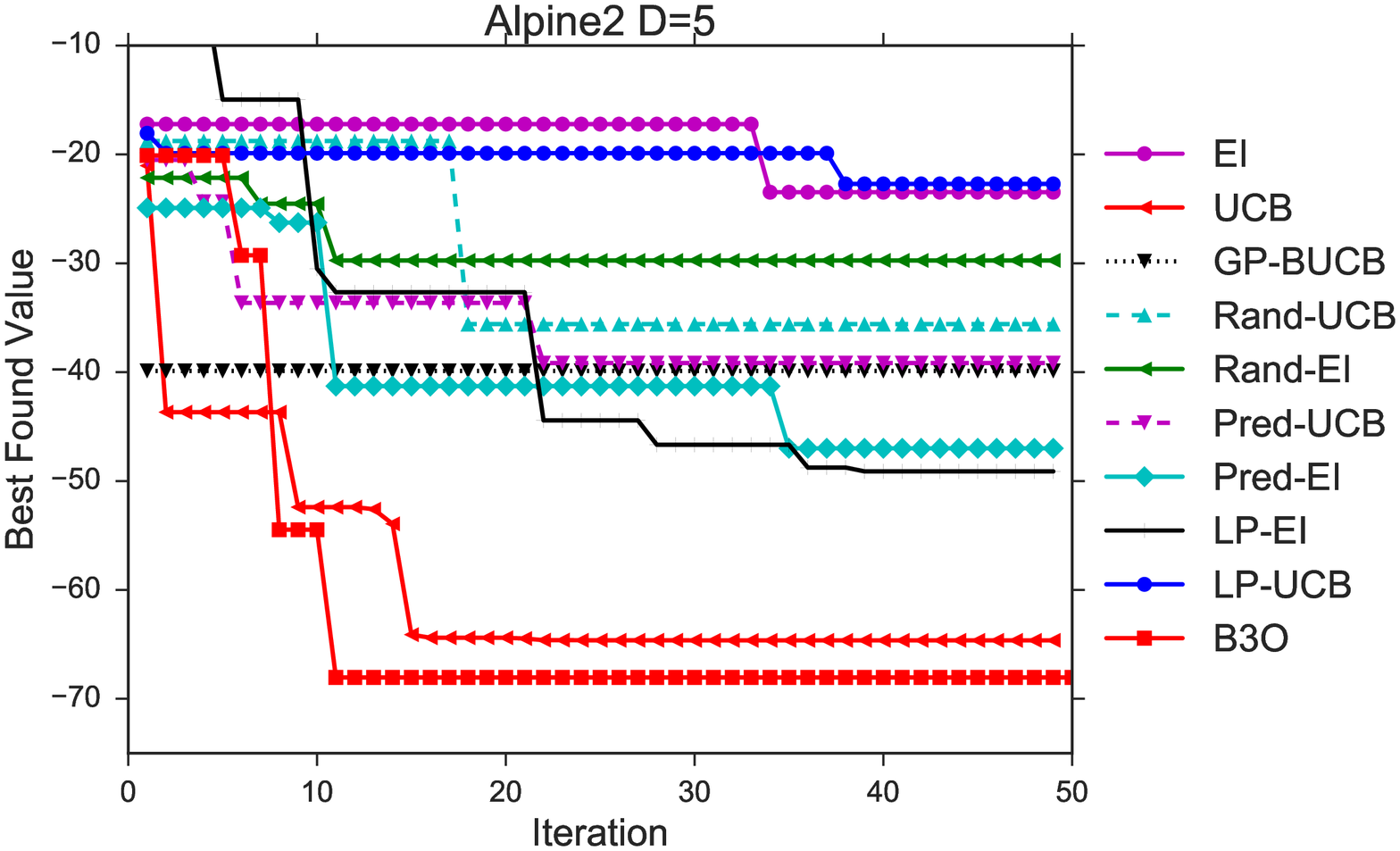}
\par\end{centering}
\begin{centering}
\includegraphics[width=0.51\columnwidth]{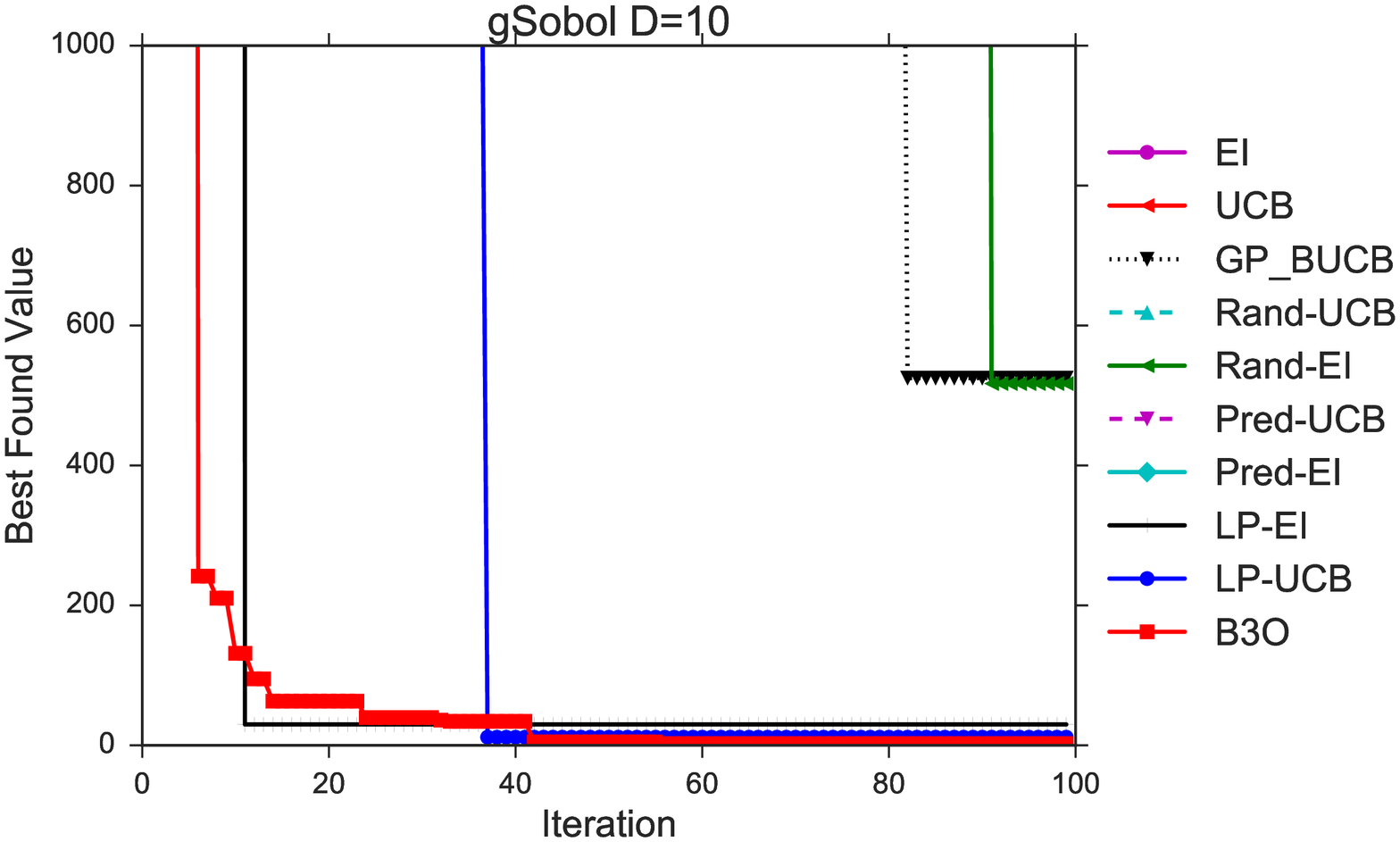}\hfill{}\includegraphics[width=0.48\columnwidth]{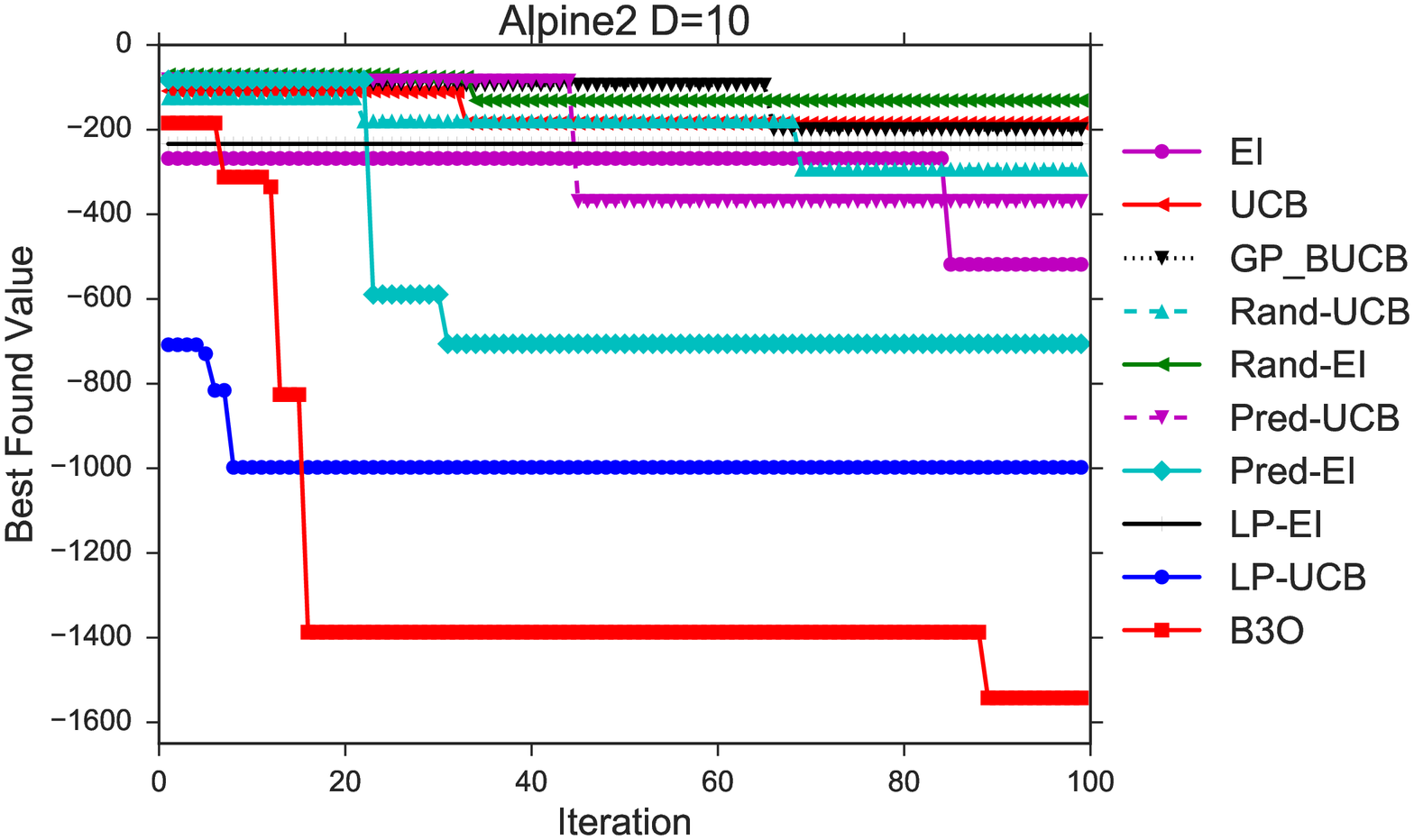}
\par\end{centering}
\caption{Best found value w.r.t. fixed number of iterations $T=10\times D$
where $D$ is the dimension. B3O is indicated in \textcolor{red}{red}\textcolor{black}{.
EI and UCB }are the sequential algorithms whilst the other baselines
are batch BO. B3O achieves the best performance for Forrester 1D,
gSobol 5D,10D and Alpine2 5D, 10D. \label{fig:Best-found-value}}
\end{figure*}

Although B3O is designed to work with any kind of acquisition function,
in this paper we use B3O with UCB since we empirically notice that
the UCB generally works better than EI for B3O. \emph{All the source
codes and data are available for reproducibility at the link}\footnote{https://github.com/ntienvu/ICDM2016\_B3O}.

\subsubsection*{Test functions and evaluation criteria}

Test functions are important to validate the performance of optimization
algorithms. There have been many benchmark functions reported in the
literature \cite{Jamil_2013Literature}. We select 5 popular benchmark
functions, see Table \ref{tab:Experimental-functions}. We consider
two major criteria for evaluation including the best-found-value $f\left(\bx_{\textrm{best}}\right)$
and the total number of evaluations $\sum_{t=0}^{T}n_{t}$.

\subsection{Comparison - best found value}

We examine the performance of B3O in finding the optimum of the chosen
benchmark functions. We demonstrate that our method can find better
optimal values (minimum) for a fixed amount of iterations $T$. We
report the best-found-value w.r.t. iterations in Fig. \ref{fig:Best-found-value}.
Our method achieves significantly better values than the baselines
in 5 over 8 functions. In particular, B3O outperforms all baselines
in Forrester $1D$, Alpine2 $5D$, gSobol $5D$, Alpine2 $10D$, gSobol
$10D$. In other three cases, B3O still provides relatively good values.
The fixed batch approaches (e.g., LP) may suffer the under-specification
(the number of specified batch size $n_{t}$ is smaller than the number
of real peaks) at some iterations and thus negatively affect the final
performance.

As expected, the batch approaches are always better than the sequential
approaches using EI and UCB because the number of evaluated points
for batch methods is much higher than the sequential. This fact also
highlights the advantage of batch BO over the standard (sequential)
BO.

\subsection{Comparison - number of evaluations}

\begin{figure}
\centering{}\includegraphics[width=0.85\columnwidth]{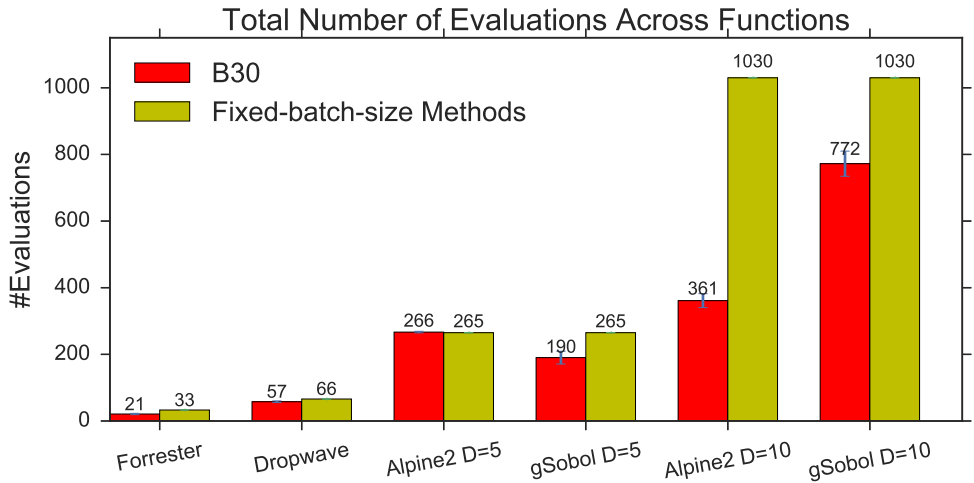}\caption{The ideal methods should require less (total) number of evaluations
$N$ for saving time and resources. Our proposed B3O always requires
less evaluations than the fix-batch size approaches.\label{fig:Comparison-number_evaluations}}
\end{figure}
The next criteria for comparison is the total number of evaluations,
$N=\sum_{t=0}^{T}n_{t}$. The requirement of Bayesian optimization
is to keep the number of evaluations to find the optimal value as
low as possible, as these evaluations can be costly. For example,
it can very expensive to perform a single experiment in material science
(to cast an alloy testing), or it takes a few days to train a deep
network on a large-scale dataset.

It is not natural to fix the number of evaluations per batch $n_{t}$,
since the number of peaks is unknown, and importantly, these peaks
will be changed after new evaluations are done. Therefore, we may
waste time and resources if we over-evaluate the number of points
than we need. In particular, after all the actual peaks are detected,
if we set the batch size $n_{t}$ large, we will get the noisy points
which are possibly close to the already detected ones due to the effect
of penalizing the peaks \cite{Gonzalez_2015Batch}. Hence, these noisy
points are not useful for evaluation. In contrast, under-evaluating
the number of necessary points also brings detrimental effects of
losing the optimal points. 

\begin{figure}
\begin{centering}
.\includegraphics[width=0.7\columnwidth]{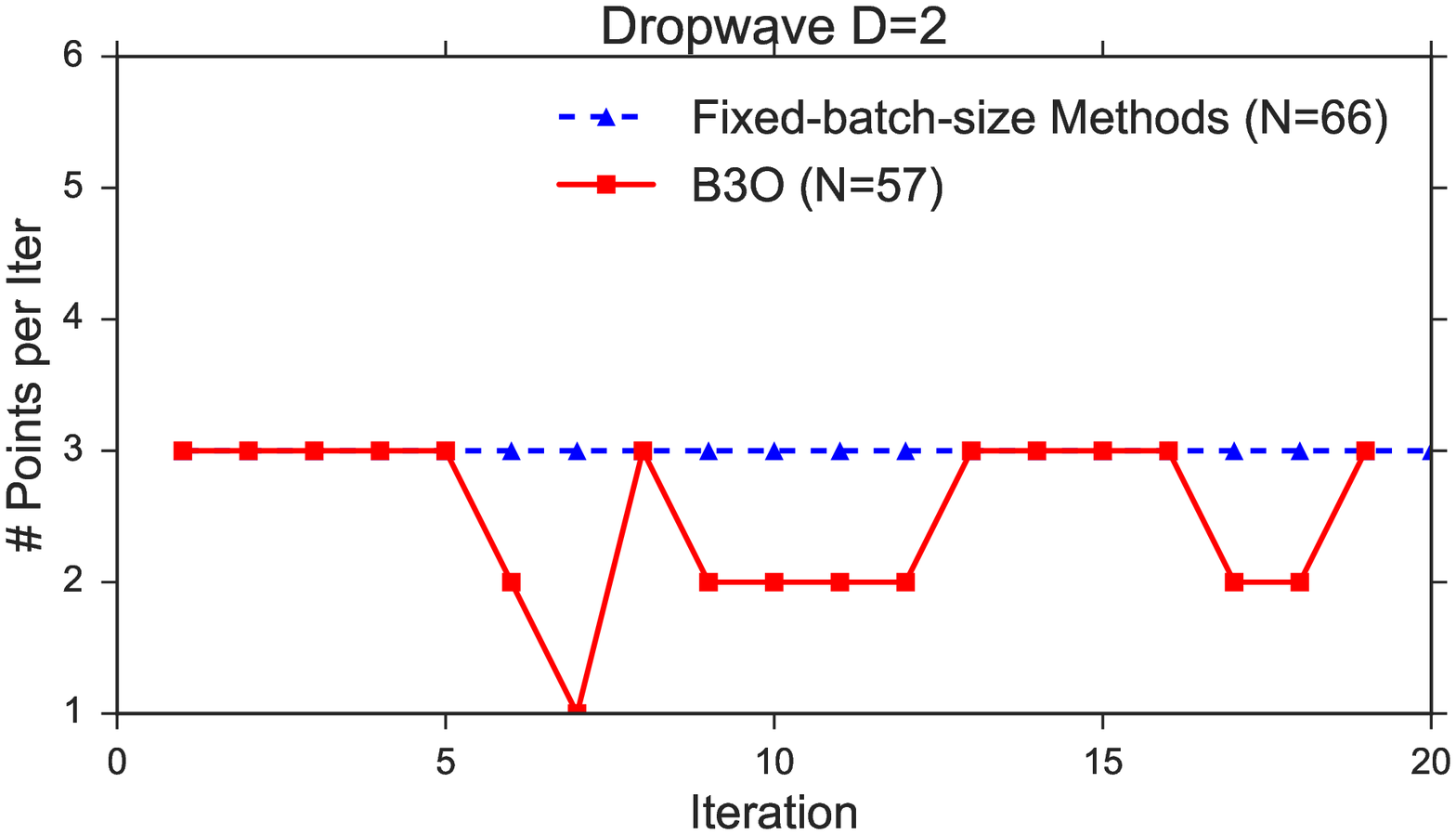}
\par\end{centering}
\vspace{5pt}

\begin{centering}
\includegraphics[width=0.7\columnwidth]{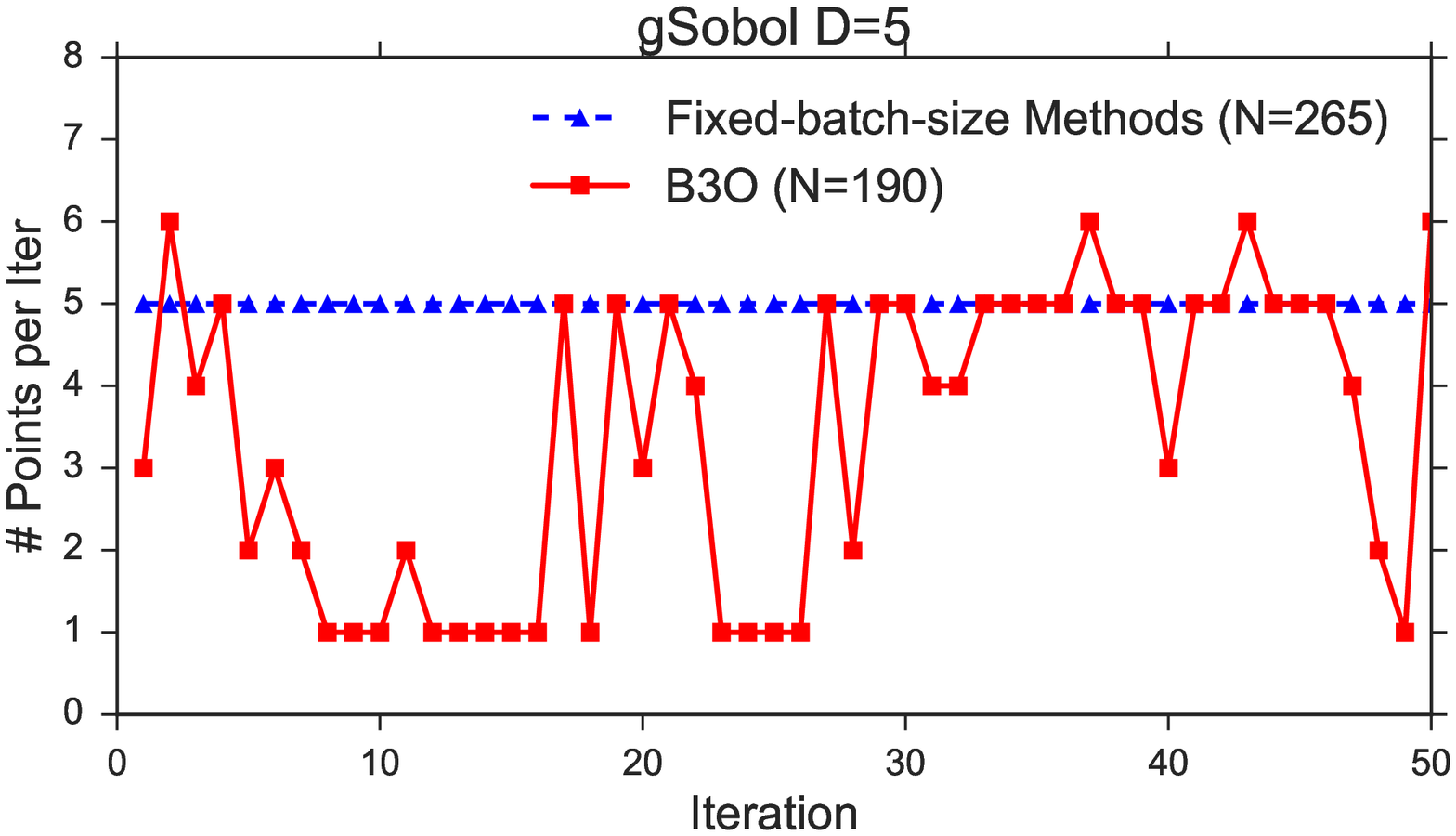}
\par\end{centering}

\caption{Number of evaluations, $n_{t}$, per iteration on Dropwave and gSobol
functions. While the existing approaches fix the batch size $n_{t}$
to a constant for all iterations, our B3O is more efficient by flexibly
creating a suitable batch size per iterations. One the one hand,
at the iteration $2$ in gSobol function, the fixed-batch-size approaches
suggest 5 points while we have 6 peaks. Because of under-specifying
the peaks, the baselines may miss the optimum. On the other hand,
at the iterations $8,9$ and $10$, the baselines suggest 5 points
while we have only a single peak. This overspecifying the peaks will
waste time and resources to evaluate at redundant points. \label{fig:Number-of-evaluation_per_iteration}}
\end{figure}
B3O automatically identifies the suitable number of peaks from the
acquisition function in each iteration, and thus does not have to
resort to a fixed batch sizes. This is efficient without suffering
any performance loss. Given a fixed number of iterations $T$, we
summarize the total number of evaluations in Fig. \ref{fig:Comparison-number_evaluations}
- our proposed approach significantly reduces the number of evaluations
as compared to the fixed-batch baselines, especially for high dimensional
settings, e.g., Alpine2 $10D$ and gSobol $10D$.

\subsection{Analysis of B3O - number of points per iteration}

We demonstrate that B3O is flexible in identifying the required number
of evaluations per batch. We study the number of estimated points
in B3O at each iteration. In particular, we record the number of points
$n_{t}$ at each iteration, see Fig. \ref{fig:Number-of-evaluation_per_iteration}
for two functions - Dropwave 2D and gSobol 5D. The estimated points
per batch in our approach is flexible and determined automatically
in this Bayesian nonparametric setting. For some iterations, B3O only
recommends one point per batch (see gSobol 5D in Fig. \ref{fig:Number-of-evaluation_per_iteration}).
This fact is essential for the homogeneous acquisition function which
contains a single peak, whereas the fixed-batch approaches may be
wasted to evaluate at some unnecessary points. Moreover, given the
fixed batch size for all iterations, other approaches may suffer the
effect of over-specification at some iterations and under-specification
at other iterations.

\subsection{Analysis of computational time \label{subsec:ComparisonComputationalTime}}

Next we compare the computational time at each iteration for the different
batch approaches. Specifically, we are interested in the CPU time
for finding the batch between different approaches. The sequential
methods and the random batch approaches are the cheapest to compute
than the other batch approaches. Thus, we do not compare with the
sequential and the random batch approaches.

Constant Liar \cite{Ginsbourger_2008Multi} and BUCB \cite{Desautels_2014Parallelizing}
take time for re-estimation of the GP when the fake observations are
added and noticeably slower when the number of data points $N$ is
large. The Local Penalization approach \cite{Gonzalez_2015Batch}
consumes a considerable amount of time to estimate the Lipschitz constant.
In addition, LP computes and maintains the penalized cost around
the visited points. The cost for penalizing and optimizing will grow
for high dimensions and/or large number of observed points. 
\begin{table*}
\begin{centering}
\par\end{centering}
\begin{centering}
\par\end{centering}
\begin{centering}
\par\end{centering}
\begin{centering}
\begin{tabular}{|c|c|c|c|c|c|c|c|}
\hline 
\rowcolor{header_color}Functions & Dim & GP-BUCB & CL-EI & CL-UCB & LP-EI & LP-UCB & B3O\tabularnewline
\hline 
\hline 
Forrester & 1D & 23.9 & 17.9 & 19.9 & 17.0  & 10.3 & \textbf{2.04}\tabularnewline
\hline 
\rowcolor{even_color}Dropwave & 2D & 81.1  & 30.4 & 37.3 & 14.3  & 10.9  & \textbf{7.34}\tabularnewline
\hline 
Hartman & 3D & 154.7  & 79.8  & 95.5 & 25.7 & 53.2 & \textbf{21.6 }\tabularnewline
\hline 
\rowcolor{even_color}Alpine2 & 5D & 573  & 277.9  & 95.5 & 183.2  & 96.0  & \textbf{58.34}\tabularnewline
\hline 
gSobol & 5D & 1360  & 599  & 500.9  & 261.7  & \textbf{61.1} & 87.35 \tabularnewline
\hline 
\rowcolor{even_color}Hartman & 6D & 1585 & 810  & 1093 & 96.4  & 235.8  & \textbf{81.85}\tabularnewline
\hline 
Alpine2 & 10D & 15,765 & 13,686 & 17,211 & 13,065 & \textbf{504.5} & 2751\tabularnewline
\hline 
\rowcolor{even_color}gSobol & 10D & 23,384 & 21,413 & 15,401 & 7,748 & \textbf{4,334} & 7,726\tabularnewline
\hline 
\end{tabular}
\par\end{centering}
\begin{centering}
\par\end{centering}
\centering{}\caption{Optimization time (sec) per iteration used by different batch BO approaches.Constant
Liar and GP-BUCB consumes the most time for updating the Gaussian
process when the fake observations are sequentially taken. Local penalization
consumes a considerable amount of time to estimate the Lipschitz constant
and maintains the penalized cost around the visited points. B3O is
the fastest approach for low dimensional functions (e.g., less than
6 dimensions). \label{tab:Running_Time}}
\end{table*}

B3O is generally competitive to LP \cite{Gonzalez_2015Batch} in terms
of computation. We run faster for low dimensional functions (e.g.,
less than 6 dimensions). However, LP tends to run faster than B3O
for 10D functions. We note that in high dimensional functions, the
bottle neck in LP is estimating the Lipschitz constant whilst for
B3O the batch generalized slice sampling is the bottle neck due to
the nature of the sampling algorithm. 

\subsection{Analysis of batch generalized slice sampling (BGSS) \label{subsec:Analysis-of-BatchSliceSampling}}

We investigate the efficiency of the BGSS presented in Section \ref{subsec:Batch-Slice-Sampler}
for drawing samples under the acquisition function. We vary the observation
dimension $D$ - $5,10,20,40,50$ - and build the acquisition function
$\alpha_{UCB}$. Then, we design two BGSS settings of size $100$
and $200$ and compare with the generalized slice sampling which
is presented in Fig. \ref{fig:Illustration-of-GeneralizedSS}. We
record two important factors including computational time and the
number of accepted samples under the curve.

\begin{figure}
\begin{centering}
\includegraphics[width=0.75\columnwidth]{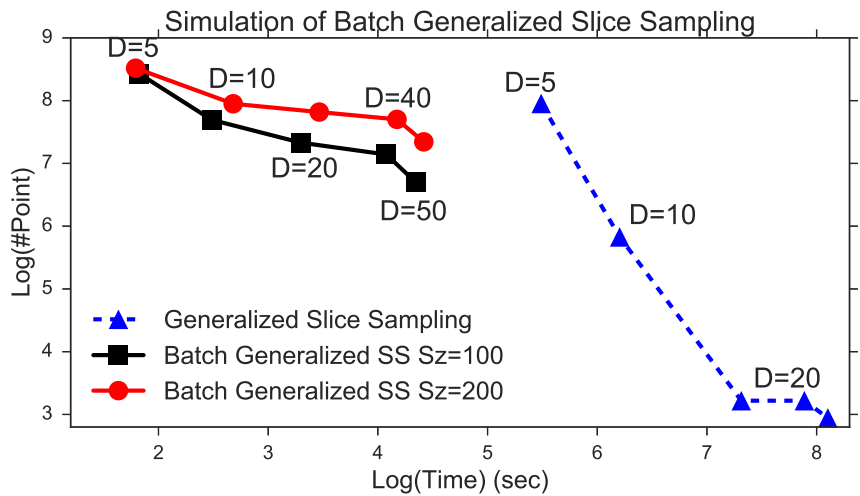}
\par\end{centering}
\caption{Simulation of Batch generalized slice sampling (BGSS) in the logarithmic
scales. Generalized slice sampling (GSS) is less efficient and slower
than the batch counterpart. Overall, our proposed BGSS can still well
handle up to $50$ dimensions. \label{fig:Simulation-of-SliceSampling}}
\end{figure}

In Fig. \ref{fig:Simulation-of-SliceSampling}, we present the simulation
results that BGSS significantly outperforms the standard slice sampling
in terms of computation (running faster) and efficiency (higher number
of accepted points). For example, BGSS of the size $200$ takes $32$
seconds to get $2485$ accepted data points at $D=10$ dimension.
These accepted points are lying under the curve and generally surrounding
the peak. Therefore, these samples are beneficial to fit the IGMM
since we are particularly interested in finding the peaks.

\subsection{Tuning hyper-parameters for machine learning algorithms}

Hyper-parameter settings greatly impact prediction accuracy of machine
learning algorithms. The traditional way of performing hyper-parameter
tuning has been to perform a grid search. In practice, however, Bayesian
optimization has been shown \cite{Bergstra_2011Algorithms,Snoek_2012Practical,Thornton_2013Auto}
to obtain better results in fewer experiments than required by a full
grid search. We employ the batch Bayesian optimization to find the
optimal hyper-parameters for support vector regression (SVR), Bayesian
nonparametric multi-label classification (BNMC), and multi layer perceptron
(MLP). 

\begin{figure}
\subfloat[Numerical comparison using RMSE, F1, Accuracy and Hardness. EI and
UCB denote for the sequential (non-batch) setting. B3O achieves the
best performances on three over four settings.\label{tab:Tuning-parameters_SVR}]{\begin{centering}
\par\end{centering}
\begin{centering}
\begin{tabular}{|c|c|c|c|c|}
\hline 
\rowcolor{header_color}Settings & \multicolumn{1}{c|}{SVR, D=3} & BNMC, D=6 & MLP, D=7 & Alloy, D=4\tabularnewline
\hline 
\rowcolor{subheader_color}Task & \multicolumn{3}{c|}{Hyper-parameter Tuning} & Experimental Design\tabularnewline
\hline 
\rowcolor{childheader_color}Evaluation & RMSE & F1 & Accuracy(\%) & Hardness\tabularnewline
\hline 
\hline 
EI & 1.935(0.01) & 0.7141(0.01) & 98.29(0.00) & 84.86(2.0)\tabularnewline
\hline 
\rowcolor{even_color}UCB & 1.937(0.01) & 0.7135(0.01) & 97.61(0.01) & 85.48(1.8)\tabularnewline
\hline 
GP-BUCB & 1.932(0.00) & 0.7189(0.01) & 98.45(0.00) & 88.23(0.2)\tabularnewline
\hline 
\rowcolor{even_color}Rand-EI & 1.933(0.00) & 0.7159(0.1) & 98.41(0.00) & 87.76(0.8)\tabularnewline
\hline 
Rand-UCB & 1.936(0.00) & 0.7149(0.1) & 98.44(0.00) & 87.87(0.9)\tabularnewline
\hline 
\rowcolor{even_color}CL-EI & 1.936(0.00) & 0.7184(0.01) & 98.44(0.00) & 86.95(1.6)\tabularnewline
\hline 
CL-UCB & 1.937(0.00) & \textbf{0.7190(0.01)} & 98.44(0.00) & 86.98(0.6)\tabularnewline
\hline 
\rowcolor{even_color}LP-EI & 1.932(0.00) & 0.7179(0.01) & 98.45(0.00) & 87.71(0.7)\tabularnewline
\hline 
LP-UCB & 1.936(0.00) & 0.7188(0.01) & 98.38(0.00) & 86.40(1.9)\tabularnewline
\hline 
\rowcolor{even_color}B3O & \textbf{1.928(0.00)} & 0.7188(0.01) & \textbf{98.47(0.00)} & \textbf{88.30(0.3)}\tabularnewline
\hline 
\end{tabular}
\par\end{centering}
\begin{centering}
\par\end{centering}
}

\subfloat[Total number of evaluations for real-world applications. B3O requires
less number of evaluations than the baselines.\label{fig:Total-=000023evaluations-RealApplications}]{\centering{}\includegraphics[width=1\columnwidth]{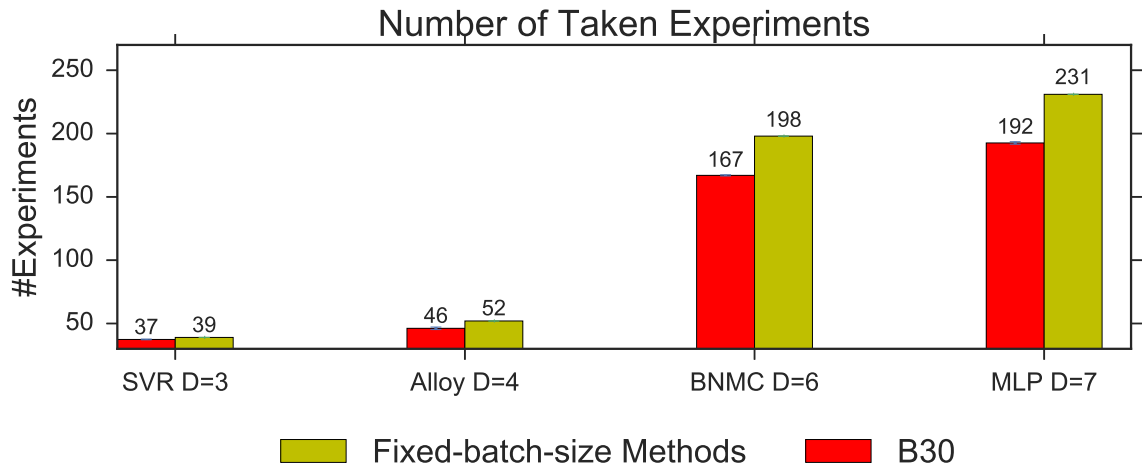}}

\caption{Performance comparison on real-world experiments. We consider tuning
hyper-parameters for three machine learning algorithm: Support vector
regression (SVR), Bayesian nonparametric multi-label classification
(BNMC), and multi layer perceptron (MLP). We also conduct the Alluminium-scandium
heat-treatment design.}
\end{figure}

\paragraph{Support Vector Regression.}

The support vector machines \cite{cortes1995support} are well-known
for classification problem. It is also applicable for regression task
as the support vector regression (SVR) model \cite{Drucker_1997Support}.
Using Abalone dataset, we consider tuning the hyper-parameters for
SVR with three parameters: $C$ (regularizer parameter), epsilon ($\epsilon$-insensitive
loss) for regression and $\gamma$ (RBF kernel function). We utilize
the Python code in sklearn package.

\paragraph{Bayesian Nonparametric Multi-label Classification.}

We select to tune the hyper-parameters for the multi-label classification
machine learning algorithm of BNMC \cite{Nguyen_2016Bayesian} on
Scene dataset using the released source code, already constructed
as a black-box function\footnote{https://github.com/ntienvu/ACML2016\_BNMC}.
BNMC uses stochastic variational inference (SVI) and stochastic gradient
descent (SGD) for learning the hidden correlation of label and feature
for multi-label prediction. In particular, we optimize 6 hyper-parameters:
Dirichlet symmetric for feature and for label, learning rate for SVI
and for SGD, truncation threshold and stick-breaking parameter. We
aim to maximize the F1 score.

\paragraph{Multi Layer Perceptron.}

For the deep learning model, we use three layers in the model and
evaluate it on the MNIST dataset. Our MLP model has 7 parameters
including 4 parameters that include the number of hidden units and
dropout rates for the first two layers and 3 parameters of learning
rate, decay, and momentum for SGD which is used to learn the MLP.
We utilize the Python code in sklearn package.

We compare B3O with baselines in Table \ref{tab:Tuning-parameters_SVR}
(columns 2-3). Our model performs better than all baselines in terms
of RMSE for SVR, F1 score for BNMC and has the highest accuracy of
$98.47\%$ for MLP. We run 10 iterations for all methods in which
the number of initial points is $9$ and the batch size $n_{t}$ is
$3$ for the fixed batch BO. The number of evaluations in B3O is the
lowest compared to other methods in this setting. Our model identifies
the unknown number of peaks in the acquisition function, then evaluate
at these points. Other fixed-batch approaches may evaluate points
which are unnecessary. Thus, B3O requires less number of evaluation
than the baselines (cf. Fig. \ref{fig:Total-=000023evaluations-RealApplications}).
Although we can run these machine learning algorithms in parallel,
each evaluation takes a significant amount of time.

\begin{figure}
\begin{centering}
\includegraphics[width=0.75\columnwidth]{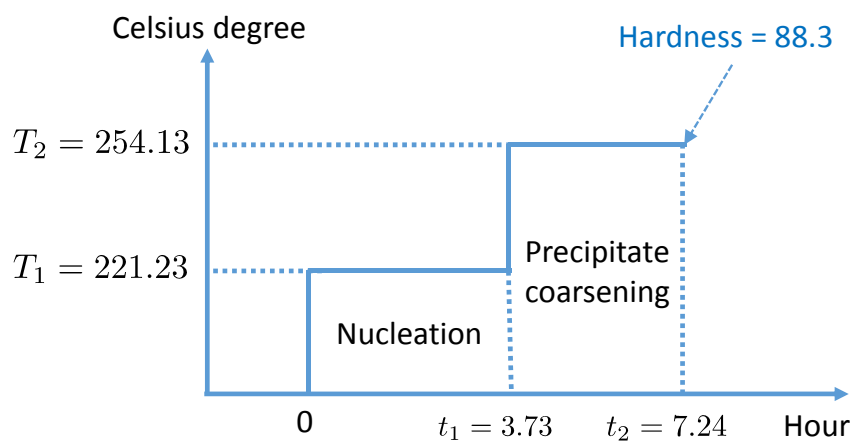}
\par\end{centering}
\caption{Aluminum-scandium hardening process. We optimize the hardness $y$
w.r.t four features $\protect\bx$ as temperatures $T_{1},T_{2}$
and times $t_{1},t_{2}$.\label{fig:Aluminium-scandium-hardening-pro}}
\end{figure}

\subsection{Bayesian optimization for experimental design}

We consider the alloy hardening process of Aluminum-scandium \cite{Wagner_1991Homogeneous}
consisting of two stages: nucleation and precipitation coarsening.
We aim to maximize the hardness for Aluminum-scandium alloys by designing
the appropriate times and temperatures for the two stages in the process.
The traditional approach is to use iterative trial and error, in
which we choose some material design (temperature and time) based
on intuition, past experience, or theoretical knowledge; and test
the material in physical experiments; and then use what we learn from
these experiments in choosing the material design to try next. This
iterative process is repeated until some combination of success or
exhaustion is reached! Since the parameter space grows exponentially,
trial and error approach is ill-affordable in terms of both time and
cost. Therefor, Bayesian optimization is a necessary choice for this
experimental design.

We use B3O for the experimental design problem above and show that
our method achieves the highest hardness using the smallest number
of experiments (see Table \ref{tab:Tuning-parameters_SVR}). The heat-treatment
profile is shown in Fig. \ref{fig:Aluminium-scandium-hardening-pro}
using the best estimated parameters (times and temperatures at two
steps) and the best hardness achieved is $88.3$. Given the times
and temperatures setting, we evaluate the hardness using the standard
kinematic KWN model used by metallurgists \cite{Kampmann_1983Kinetics}
.

\section{Conclusion}

We have introduced a novel approach for batch Bayesian optimization.
The proposed approach of B3O can identify the suitable batch size
at each iteration that most existing approaches in batch BO are unable
to do. We have presented the batch generalized slice sampling for
drawing samples under the acquisition function. We perform extensive
experiments on finding the optimal solution for 8 synthetics functions
and evaluate the performance further on 4 real-world tasks. The experimental
results highlight the ability of B3O in finding the optimum whilst
requiring fewer evaluations than the baselines.

\section{Acknowledgment }

We thank Dr Paul G Sanders and Kyle J Deane for the real-world case
study of Aluminum-scandium hardening process.

\bibliographystyle{theapa}
\bibliography{P:/03.Research/vunguyen}

\end{document}